\documentclass{article}
\usepackage{latexsym, amssymb, amsfonts}
\usepackage[dvips]{graphicx}
\usepackage{fancyvrb}

\title{`Computing' as Information Compression by Multiple Alignment, Unification and Search\protect\footnote{{\em Journal of Universal Computer Science} 5(11), 776--815, 1999.}}

\author{J Gerard Wolff\\
\small (University of Wales, Bangor, UK\\
gerry@sees.bangor.ac.uk)}

\begin{document}

\maketitle

\begin{abstract}
This paper argues that the operations of a `Universal Turing Machine' (UTM) and equivalent mechanisms such as the `Post Canonical System' (PCS) - which are widely accepted as definitions of the concept of `computing' - may be interpreted as {\it information compression by multiple alignment, unification and search} (ICMAUS).

The motivation for this interpretation is that it suggests ways in which the UTM/PCS model may be augmented in a proposed new computing system designed to exploit the ICMAUS principles as fully as possible. The provision of a relatively sophisticated search mechanism in the proposed `SP' system appears to open the door to the {\it integration} and {\it simplification} of a range of functions including unsupervised inductive learning, best-match pattern recognition and information retrieval, probabilistic reasoning, planning and problem solving, and others. Detailed consideration of how the ICMAUS principles may be applied to these functions is outside the scope of this article but relevant sources are cited in this article.
\end{abstract}


{\bf Category:} SD F.0

\section{Introduction}\label{INTRODUCTION}

For about 60 years, it has been widely accepted that the intuitive concept of `computing' may be defined in terms of the operations of a `Universal Turing Machine' (UTM, \cite{r19}) and concepts which are recognised as equivalent such as the `Lambda Calculus' (Church and Kleene, see \cite{r14}), `Recursive Function' \cite{r7}, `Normal Algorithm' \cite{r9} and Post's `Canonical System' (PCS, \cite{r11}). These and related ideas have been phenomenally successful, providing the theoretical basis for the design of most modern digital computers.

However, notwithstanding Turing's own vision of the possibility of artificial intelligence \cite{r20}, the UTM and the PCS and equivalent models of computing operate in a rigid `clockwork' manner which appears far removed from the fluidity of human intelligence. Likewise, ordinary digital computers, without special programming, are notoriously lacking in human-like flexibility and intelligence.

Because the UTM and equivalent models, and a conventional computer in `raw' state, have so little functionality and so little intelligence, it is necessary for practical purposes to create programs which provide the missing capabilities, as far as these are understood. Although our understanding is still very patchy, there is increasing success with AI applications in achieving human-like capabilities such as speech recognition, the ability to play games like chess, go or bridge, the ability to recognise visual patterns and objects in a flexible manner, and others.

Since computers were first invented, much effort has been devoted to the development of programs and associated concepts across a wide range of applications. Many areas of application are now quite well served by specialised programs. However:

\begin{itemize}
\item Notwithstanding the development of `suites' of programs intended to work together, there are still many problems of integration between different kinds of application.

\item Similar concepts have emerged in diverse forms. For example, the concept of `heuristic search' manifests itself in the form of `hill climbing' (or `descent'), `genetic algorithm', `simulated annealing', `dynamic programming', `neural computation', and others. Without normally being recognised as such, heuristic methods also appear in `standard' computational techniques such as the Newton-Raphson method for finding square roots, dynamic programming and techniques for matching patterns and retrieving information.

\item The diversity of programs and concepts is collectively, very complex. There is a need for rationalisation and simplification across diverse areas.
\end{itemize}

\subsection{New developments}

These considerations have been the main motivation for a programme of research (\cite{r35} to \cite{r21a}) seeking new insights into the nature of computing, and aiming to develop a `new generation' computing system based on these ideas.

If we can identify the elements which are shared by diverse applications and abstract them from the elements which are truly specific for each application, this may help us to achieve an overall simplification of computing systems and better integration across applications. It can mean less programming - and thus less cost - and there should be less opportunity to introduce errors - yielding gains in quality.

As we shall see (Section \ref{SP-THEORY}), the central idea in this research is the conjecture that ``all kinds of computing and formal reasoning may usefully be understood as {\it information compression by multiple alignment, unification and search} (ICMAUS)''. This is still only a conjecture and may turn out to be wrong. But there is already an accumulation of supporting evidence described and discussed in sources cited in Section \ref{SCOPE}.

The use of the word ``theory'' for these proposals (Section \ref{SP-THEORY}) may suggest that they are intended to displace established theories of computing. It is more accurate to say that the proposals are a framework of ideas which provide an interpretation for established models of computing and which suggest ways in which those models may be augmented to provide additional functionality which is generic for many applications.

As a test-bed for these ideas and a vehicle for demonstrations, a software model has been developed, SP61, running on a conventional computer. This is described briefly in Section \ref{WORKING-MODEL} and more fully in sources cited there. All the alignments shown in this paper are output from the model.

\subsection{Scope and orientation of this paper}\label{SCOPE}

Regarding the primary motivation for these developments - to promote simplicity and integration in computing - no attempt will be made in this paper to demonstrate the potential of these ideas. Readers are referred to the following sources describing how the ideas may apply in diverse areas of computing:\footnote{The papers before 1996 were written at a stage before the `multiple alignment' insights were achieved (see Section \ref{SP-THEORY}, below).}

\begin{itemize}
\item Background thinking (the pervasive nature of information compression in human cognition and also in artificial computing): \cite{r33}.

\item Unsupervised inductive learning:

\begin{itemize}
	\item The entire programme of research is based on a previous programme of research into unsupervised inductive learning of language (\cite{r36, r37} and earlier articles cited in these two articles).

	\item A demonstration of how an early form of the framework of ideas may be applied to learning is presented in \cite{r29}.

	\item Integration of unsupervised inductive learning and different kinds of reasoning: \cite{r27}. 
\end{itemize}

\item Best match information retrieval: \cite{r21a, r30}.

\item The design and execution of software: \cite{r31}.

\item \sloppy The potential range of applications of the SP concepts: \cite{r28, r29}.

\item Causality: \cite{r26}.

\item Parsing of natural language and representing the structure of natural language in the ICMAUS framework \cite{r25, r25a, r25b}.

\item Probabilistic reasoning in the ICMAUS framework \cite{r21a, r22, r23, r24}, including:

\begin{itemize}
\item Representing class hierarchies and inheritance of attributes.

\item Best-match pattern recognition and information retrieval.

\item Probabilistic `deductive' reasoning.

\item Probabilistic abductive reasoning.

\item Reasoning with probabilistic decision networks and decision trees.

\item Reasoning with probabilistic if-then rules.

\item Reasoning with default values and nonmonotonic reasoning.

\item Solving geometric analogy problems.

\item Modelling the functioning of a Bayesian network.
\end{itemize}
\end{itemize}

The primary purpose of this paper is to describe the SP theory in outline, to show how it may provide an interpretive framework for the UTM and PCS, and to describe how those models may be augmented for greater generic functionality.

Apart from Appendix \ref{APPENDIX}, the presentation is relatively informal with examples to make the ideas clear. No attempt has been made to present a formal proof that the ICMAUS framework may model any UTM/PCS, largely because a detailed proof would take far more space than is available here.

\subsection{Presentation}

In what follows:

\begin{itemize}
\item Section \ref{SP-THEORY} describes, in outline, the SP theory as it has been developed to date, together with some brief remarks about working models of the system.

\item Section \ref{UTM-PCS} describes, briefly, the concept of a UTM and, more fully, the equivalent concept of a PCS.

\item Section \ref{ICMAUS-AND-PCS} tries to show how the operation of a PCS (and, consequently, the operation of a UTM) may interpreted in terms of ICMAUS concepts. 

\item Section \ref{AUGMENTATION} considers how established models of computing may, with advantage, be augmented.

\item Section \ref{CONCLUSION} contains concluding remarks.

\item Appendix \ref{APPENDIX} defines the key terms used in this article.
\end{itemize}

\section{The SP theory}\label{SP-THEORY}

As already indicated, the central idea in the `SP' theory is the conjecture that:

\begin{quote}
{\it All kinds of computing and formal reasoning may usefully be understood as information compression by multiple alignment,}\footnote{In earlier publications, the conjecture was expressed with the words `pattern matching' in place of the words `multiple alignment'. However, in the last few years, it has seemed appropriate to focus more narrowly on multiple alignment (which is one form of pattern matching) as the most promising framework (with unification and search) for the intended integration of concepts.} {\it unification and search.}
\end{quote}

Information compression (IC) may be interpreted as a process of maximising {\it Simplicity} in information (by reducing redundancy) whilst retaining as much as possible of its non-redundant descriptive {\it Power}. Hence the mnemonic `SP' which has been applied to these ideas.

In this programme of research:

\begin{itemize}
\item The term {\it multiple alignment} (MA) has a meaning which is related to but distinct from its meaning in bio-informatics (where the term is normally used). The meaning of the term in this programme of research is explained below (Section \ref{MULTIPLE-ALIGNMENT}).

\item The term {\it unification} means a simple merging of matching symbols and sequences of symbols, a meaning which is related to but simpler than the concept of `unification' in logic (where it means the assignment of values to variables in two structures so that the two structures become the same).

	As explained more fully below (Section \ref{ROLE-OF-UNIFICATION}), the unification of matching {\it patterns}\footnote{The meaning of the term {\it pattern} in this research is defined in Appendix \ref{APPENDIX}. It includes the meaning of the term {\it sequence} but it is more general because it includes arrays of symbols in two or more dimensions and it includes sequences (and other arrays) which are discontinuous. However, in most contexts in this article, the term {\it pattern} may be read as synonymous with the term {\it sequence}.} is the core of the mechanism by which information is compressed. Conjecturally, it is the {\it sole} mechanism by which information may be compressed - meaning that {\it all} other mechanisms may be seen to contain unification at a fundamental level. More research is needed to establish whether or not this conjecture is true.

\item In this context, the term {\it search} means the systematic exploration of the abstract space of possible multiple alignments. This can mean exhaustive search of the entire space of possible alignments but, as amplified below, for most practical problems, there must be some kind of {\it constraint} which reduces the size of the abstract space being searched.

\item As we shall see (Section \ref{OTHER-EXAMPLES}), the conjecture about the nature of computing, above, depends on at least one key assumption about how IC is measured. It also needs to be qualified by the observation that, to be practical, the process of searching for MAs that yield `good' IC, it is sometimes necessary to examine MAs yielding ICs which are zero or negative (see Section \ref{ESCAPING}).
\end{itemize}

\subsection{Minimum Length Encoding and unsupervised inductive learning}\label{MLE-LEARNING}

\sloppy The SP theory incorporates one version of the principle of `Minimum Length Encoding' (MLE)\footnote{MLE is an umbrella term for concepts of `Minimum Message Length' encoding (MML) and the closely related `Minimum Description Length' encoding (MDL).} from the field of inductive inference (see \cite{r16, r21, r13, r8}). MLE is itself a version of `Occam's Razor'.

The key idea in MLE is that, {\it \bf in measuring compression, we must measure the size of the raw data in its encoded form together with the size of the code patterns which are needed to create the encoding}. Of course, if one or other of these components is constant (as is the case with all the examples considered in this article) then, for purposes of comparison, it is only necessary to measure the one which varies.

If both components can vary then, if either of them is omitted, counter-intuitive anomalies can arise. For example:

\begin{itemize}
\item We may create a code pattern comprising the entire contents of the works of Shakespeare with a one-bit label. Given that code pattern, we can compress the complete works of Shakespeare into one bit.

\item Alternatively, we can create a set of code patterns comprising the letters and other characters used to print the text of Shakespeare. In this case, the set of code patterns is very small but encoding the works of Shakespeare using these patterns gives a result which is the same size as the original!
\end{itemize}

Clearly, any realistic compression of the Shakespeare text falls somewhere between these two extremes. Minimising the overall size of the set of code patterns and the encoded text seems to produce the required balance.

In connection with the discovery and development of these ideas, \cite{r17} writes:

\begin{quote}
``{\it I was trying to find an algorithm for the discovery of the `best' grammar for a given set of acceptable sentences. One of the things I sought was: Given a set of positive cases of acceptable sentences and several grammars, any of which is able to generate all the sentences, what goodness of fit criterion should be used? It is clear that the} `ad hoc {\it grammar', that lists all of the sentences in the corpus, fits perfectly. The `promiscuous grammar' that accepts any conceivable sentence, also fits perfectly. The first grammar has a long description; the second has a short description. It seemed that some grammar half-way between these, was `correct' - but what criterion should be used?}''
\end{quote}

As indicated above, the key seems to be to minimise the overall size of the `grammar' (repository of code patterns) together with the text which is encoded in terms of the grammar.

Similar ideas have been developed in computer models designed to test whether or not `statistical' mechanisms might provide an explanation of some aspects of what children can apparently achieve when they learn their first language (see \cite{r36, r37} and earlier papers referenced there):

\begin{itemize}
\item There is evidence that children can learn the segmental structure of language (words, phrases, sentences) without needing explicit markers at the beginnings and ends of these structures or explicit teaching.

\item In a similar way, they appear to be able to `bootstrap' their knowledge of word classes and other disjunctive categories in language without the aid of a teacher or other assistance.

\item Apparently without the need for external supervision or correction, children can learn to distinguish `correct' generalisations of grammatical rules from `incorrect' overgeneralisations despite the fact that, by definition, both kinds of generalisation have zero frequency in the child's experience.
\end{itemize}

The models that have been developed (described in the sources cited above) appear to provide solutions to these problems and, as a by-product of the way they work and without {\it ad hoc} provision, exhibit several features in their learning which correspond remarkably well with a range of phenomena observed when children learn their first language. This empirical evidence is reviewed in \cite{r36}.

\subsection{Multiple alignment}\label{MULTIPLE-ALIGNMENT}

The term {\it multiple alignment} is normally associated with the computational analysis of (symbolic representations of) sequences of DNA bases or sequences of amino acid residues as part of the process of elucidating the structure, functions or evolution of the corresponding molecules.

The aim of the computation is to find one or more {\it alignments} which are, in some sense, `good'. What `good' can mean is discussed briefly below (Section \ref{EVALUATION-OF-ALIGNMENTS}).

An {\it alignment} is an arrangement of two or more sequences of atomic symbols so that, by judicious `stretching' of sequences where appropriate, symbols that match each other are arranged in vertical columns.

As a matter of presentation, it is sometimes convenient to allow some columns to contain symbols that do not match each other and in this case vertical lines between symbols are used to mark where there are positive matches ({\it hits}) between the symbols. An example of an alignment of DNA sequences which uses this second convention is shown in Figure \ref{DNA}.

\begin{figure}
\centering
\begin{BVerbatim}
  G G A     G     C A G G G A G G A     T G     G   G G A
  | | |     |     | | | | | | | | |     | |     |   | | |
  G G | G   G C C C A G G G A G G A     | G G C G   G G A
  | | |     | | | | | | | | | | | |     | |     |   | | |
A | G A C T G C C C A G G G | G G | G C T G     G A | G A
  | | |           | | | | | | | | |   |   |     |   | | |
  G G A A         | A G G G A G G A   | A G     G   G G A
  | |   |         | | | | | | | |     |   |     |   | | |
  G G C A         C A G G G A G G     C   G     G   G G A
\end{BVerbatim}
\caption{\small A `good' alignment amongst five DNA sequences.}
\label{DNA}
\normalsize
\end{figure}

\subsubsection{Generalisation of the concept of multiple alignment}\label{GENERALISATION-MA}

In this research, the concept of MA has been generalised in the following way:

\begin{enumerate}
\item One (or more) of the sequences of symbols to be aligned has a special status and is designated as `New'.

\item All other sequences are designated as `Old'.

\item Alignments are evaluated in terms of the compression of New that can be achieved by encoding New in terms of sequences in Old, with or without the hierarchical encoding of tags mentioned in Section \ref{ENCODING}, below. A detailed description of how this can be done is presented in \cite{r22, r25a} with a somewhat less detailed description in \cite{r21a}.

\item An implication of this way of framing the alignment problem is that, by contrast with `multiple alignment' as normally understood in bio-informatics, any given sequence in Old may appear two or more times in any one alignment and may therefore be aligned with itself (with the obvious restriction that any one instance of a symbol may not be aligned with itself).
\end{enumerate}

Regarding the last point, the kind of MA shown in Figure \ref{DNA} can obviously include two or more {\it copies} of a given sequence in any one alignment. It is unlikely that anyone would bother to do this in practice because it would simply lead to the trivial alignment of each symbol in one copy with the corresponding symbol or symbols in the one or more other copies.

What is proposed for the ICMAUS framework is different: any {\it one} sequence may appear two or more times in an alignment. Each appearance is just that, it is an {\it appearance} of {\it one} sequence, it is {\it not} a duplicate {\it copy} of a pattern.

Since each appearance of a pattern represents the same pattern, it makes no sense to recognise a hit between a symbol from one appearance and the corresponding symbol in another appearance - because this is simply matching one instance of the symbol with itself. Any such match is spurious and must be forbidden. However, it is entirely possible to recognise a hit between a symbol from one pattern and a {\it different} symbol from the same pattern. Examples will be seen later.

\subsubsection{A simple example}\label{SIMPLE-EXAMPLE}

To give the reader a flavour of the concept of multiple alignment as it has been developed in this research, a very simple example (from \cite{r25}) is presented here showing how parsing may be understood in these terms.

Figure \ref{GRAMMAR-PATTERNS} shows a `grammar' for a fragment of English syntax presented as {\it patterns} as defined in Section A1. This particular example models a context-free phrase-structure grammar (CF-PSG) except that there is no rewrite arrow and each pattern has a number (shown in brackets on the right) which is its notional frequency of occurrence in a parsing of a sample of English.\footnote{Despite the similarity, in this example, with a CF-PSG, this mode of representing syntax appears to have the kind of context-sensitive expressive power which is needed to represent the syntax of natural languages and which is missing from CF-PSGs (see \cite{r25, r25a, r25b}).} 

\begin{figure}
\centering
\begin{BVerbatim}
S N #N V #V #S (1000)
N 0 j o h n #N (300)
N 1 s u s a n #N (700)
V 0 w a l k s #V (650)
V 1 r u n s #V (350)
\end{BVerbatim}
\caption{\small A simple grammar written as patterns of symbols. For each pattern, there is a number in brackets on the right representing the frequency of occurrence of the pattern in parsing of a notional sample of the language.}
\label{GRAMMAR-PATTERNS}
\normalsize
\end{figure}

Figure \ref{PARSING} shows an MA which may be interpreted as a parsing of the sentence `j o h n r u n s' in terms of the grammar in Figure \ref{GRAMMAR-PATTERNS}. In parsings like this, the sentence to be parsed plays the role of New and the grammar used for parsing is treated as Old. By convention in these kinds of MA, the pattern which is New occupies the first row of the alignment and relevant patterns from Old are shown in the second and subsequent rows, one pattern per row.

\begin{figure}
\centering
\begin{BVerbatim}
0       j o h n        r u n s       0
        | | | |        | | | |       
1   N 0 j o h n #N     | | | |       1
    |           |      | | | |       
2 S N           #N V   | | | | #V #S 2
                   |   | | | | |     
3                  V 1 r u n s #V    3
\end{BVerbatim}
\caption{\small Parsing of the sentence `j o h n r u n s' as an alignment amongst sequences representing the sentence and relevant rules in the grammar
in Figure \ref{GRAMMAR-PATTERNS}.}
\label{PARSING}
\normalsize
\end{figure}

\subsection{Evaluation of alignments in terms of information compression}\label{EVALUATION-OF-ALIGNMENTS}

Intuitively, a good alignment is one which has many {\it hits} (positive matches between symbols), few {\it gaps} (sequences of one or more symbols which are not part of any hit) and, where there are gaps, they should be as short as possible. It is possible to use measures like these directly in computer programs for finding good MAs and, indeed, they commonly are. However, our confidence in the validity of measures like these may be increased if they can be placed within a broader theoretical framework.

In keeping with the remarks about IC, above, it seems that IC, and related concepts of probability,\footnote{Variations in the probability of symbols and patterns has a key role in compression methods such as Huffman coding or Shannon-Fano-Elias coding (see \cite{r5}). Conversely, measures of information content after compression can be used to calculate probabilities (see \cite{r17, r21a, r22, r23, r24}).} may provide that framework. Work on the evaluation of MAs in this tradition includes \cite{r12, r6, r3, r1, r1a, r30}.

In the present proposals, MAs are also evaluated in terms of IC. However, owing to the way in which the concept of multiple alignment has been generalised (Section \ref{GENERALISATION-MA}, above), the way in which alignments are evaluated in terms of IC, described in outline next, is somewhat different from methods used elsewhere.

\subsubsection{Encoding New in terms of codes for patterns in Old}\label{ENCODING}

There is insufficient space to explain in detail how alignments are evaluated in terms of IC in the present scheme. Detailed explanation and examples showing how this can be done with multiple alignments can be found in \cite{r25a, r22} with somewhat less detailed accounts in \cite{r25, r21a}. This subsection describes the method in outline.

In Figure \ref{PARSING}, `j o h n' in New is aligned with `N 0 j o h n \#N' in Old. It should be intuitively clear that we can encode this part of New using the `code' symbols `N 0 \#N'. Likewise `r u n s' may be encoded as `V 1 \#V'. Thus the whole sentence may be encoded as `N 0 \#N V 1 \#V' with a modest reduction from 8 symbols to 6 (assuming, for the sake of simplicity in this example, that all symbols are encoded with the same number of bits).

However, the pattern `N 0 \#N V 1 \#V' contains the subsequence `N \#N V \#V'. And this subsequence is contained within the pattern `S N \#N V \#V \#S' - which is one of the patterns in Old. This means that the subsequence `N \#N V \#V' may be encoded as `S \#S'. Thus the sequence `N 0 \#N V 1 \#V' may be reduced to `S 0 1 \#S', preserving the symbols `0' and `1' to encode `j o h n' and `r u n s' respectively.

This encoding at two `levels' (the sentence level and the word level) has the effect of reducing the original 8 symbols to 4. It should be clear that, with more complex examples, the same principles may be applied recursively through any number of `levels'.

Regarding the MLE principle (Section \ref{MLE-LEARNING}) that measures of compression should include both the size of the `grammar' used for coding and the size of the data after encoding, this is true in the context of systems that learn - where the size of the grammar is changing. As was noted previously, this need not apply in examples like the ones considered in this article, where, in each example, the size of Old remains constant. In these cases, for purposes of comparison, we may ignore the size of Old.

\subsubsection{Taking account of the probabilities of patterns and symbols}

To present the method simply, this outline has assumed that every symbol is encoded with the same number of bits as every other symbol. However, the method used in the SP61 model (see Section \ref{WORKING-MODEL}, below) calculates the encoding cost of each symbol individually, taking account of its frequency of occurrence using a formula based on the Shannon-Fano-Elias (S-F-E) method (See \cite{r5}). Of course, encoding costs calculated in this way are totally unrelated to the numbers of characters used in the names of symbols. Names of symbols are used purely for the sake of human comprehension and readability.

The sizes of codes assigned to patterns may, in a similar way, be related to their frequencies of occurrence using the S-F-E method or the Huffman method or something similar. However, for reasons which would take too much space to explain here, the SP61 model determines the size of the code for each pattern from the sizes of the symbols which are used in the code for the given pattern. This yields code sizes which are roughly in line with the principle that rare patterns should have longer codes than common patterns but the result is not as precise as with Huffman and similar methods.

\subsubsection{The role of unification in information compression}\label{ROLE-OF-UNIFICATION}

It may seem too natural and `obvious' to command much attention but the essence of the compression just described is the merging or {\it unification} of two or more sequences of symbols to make one. At the lowest level, `j o h n' from New is, in effect, unified with `N 0 j o h n \#N' from Old. At the next level, `N \#N V \#V' from two of the patterns in Old is unified with `S N \#N V \#V \#S', also from Old.

It is this unification of matching patterns, with or without the use of `codes' or `tags' to show where information has been deleted, which provides the foundation of all the simpler `standard' methods for IC (see \cite{r18}) and, by conjecture, all other methods as well.

In the ICMAUS scheme, the touchstone of whether the necessary unifications can be applied is whether or not an alignment can be `projected' into a single sequence without any ambiguity about the left-to-right sequence of the symbols. In the example shown in Figure \ref{PARSING}, the alignment can be projected into a single sequence, thus:

\begin{center}
\begin{BVerbatim}
S N 0 j o h n #N V 1 r u n s #V #S.
\end{BVerbatim}
\end{center}

In this projection, each symbol corresponds to a column in the alignment and, in each column containing two or more symbols, the symbols have been unified to make one. Every column has a place in the alignment which ensures that its left-right position relative to other columns is not ambiguous.

\subsubsection{`Mismatches' in alignments}

By contrast with the example just given, the two `nonsense' alignments shown here:

\begin{center}
\begin{BVerbatim}
a   b       a b
|   |       | |
a x b  and  a b x
|   |       | |
a y b       a b y
\end{BVerbatim}
\end{center}

\noindent exhibit ambiguity about the relative left-to-right positions of `x' and `y'. This means that neither of the alignments can be projected into a single sequence without making arbitrary choices, not supported by the data, about whether `x' should precede `y' or {\it vice versa}.

This kind of ambiguity in an alignment is described, in this research, as a `mismatch' of symbols. In the ICMAUS scheme as it has been developed to date (and the SP61 model in which the scheme has been expressed) any alignment containing one or more mismatches between symbols in Old is discarded. It is possible, of course, to devise some scheme for resolving this kind of conflict using some kind of measure of `strength' or `support' for each of the two alternatives. At present, however, the theory in its current form does not provide any motive for this kind of manoevre and there seems not to be any need for it. Thus, pending new insights which might dictate a different course of action, any alignment of this kind is simply discarded.

In the ICMAUS scheme as currently conceived and also in the SP61 model, mismatches between symbols in New and symbols in Old are taken to mean simply that the offending part of New cannot be encoded in terms of the given alignment. Alignments with this kind of mismatch are permitted in the ICMAUS scheme and seem to be necessary in most parsing applications so that larger alignments encoding relatively large parts of New can be built from smaller ones which encode relatively small sections of New.

\subsubsection{Information compression and Algorithmic Information Theory}

In Algorithmic Information Theory (see, for example \cite{r8}), a body of information is deemed to be random if no computer program can be found to generate the information which is smaller than that body of information. This elegant approach to defining randomness (which has advantages over earlier concepts of randomness) suggests that information compression should be defined in terms of relative sizes of computer programs.

The types of encodings described above may not look much like computer programs but there are remarkably close parallels between the devices used in structuring computer programs and the devices used for information compression. It would take too much space to discuss this here but readers may find a relatively full discussion in \cite{r33}.

\subsection{Search}

It is generally understood that, in multiple alignment problems, the abstract space of possible alignments (and corresponding unifications) is normally too large to be searched exhaustively. For example, the size of the search space for alignments of one sequence (A) of length $m$ with another sequence (B) of length $n$ is:

\[\mathit{\Psi} = \sum_{i = 1}^{m}\sum_{j = 1}^{n}(a_i \times b_j)\]

\noindent where $a_i$ is the number of subsequences of length $i$ within A and $b_j$ is the number of subsequences of length $j$ in B, and $a_i$ is calculated as:

\[a_i = \frac{m!}{i!(m - i)!}\]

\noindent and likewise for $b_j$, substituting $n$ for $m$. Even in simple cases like these, the size of the search space grows exponentially with the size of the patterns. In general, search spaces are astronomically large except when patterns are very few and very short.

\subsubsection{Constraints on search}\label{CONSTRAINTS}

Because the entire search space for an MA is normally so very large, it is necessary for practical purposes to constrain the search in some way so that only a part of the search space is examined. In broad terms, this can be done in two different ways (which may be applied singly or together):

\begin{itemize}
\item More or less arbitrary constraints may be applied. For example, a limit may be placed on the maximum size of any gap (sequence of unmatched symbols) in any alignment. All other alignments are excluded from consideration. With this kind of constraint, certain parts of the search space are ruled out {\it a priori} and can never be reached.

\item The second kind of constraint is applied in `heuristic' techniques like `hill climbing' (`descent'), `beam search', `genetic algorithm', `simulated annealing' and so on. These techniques use some kind of measure of `goodness' to narrow the space which is to be searched in successive stages. For this reason, they are sometimes known generically as `metrics-guided search'. With heuristic techniques, the space which is to be searched can be dramatically reduced but no part of the total search space is ruled out {\it a priori}.
\end{itemize}

With constraints like these, there is normally a trade-off between the `quality' of the best solutions found and the thoroughness of the search. With these kinds of technique, ideal solutions can be guaranteed only when examples are {\it very} small. However, with examples of realistic size, narrowing the search space as described above can yield good approximate solutions without excessive computational demands. For many practical applications, these approximate solutions are quite acceptable.

\subsubsection{Escaping from `local peaks'}\label{ESCAPING}

As just noted, a recognised feature of heuristic search techniques like hill climbing is that the search process may reach solutions to the search problem which are relatively good but are not the best that may be found. With all but the very smallest examples, it is not possible to prove that a given solution is the best possible solution. With every solution, there is normally the possibility that a better solution may exist elsewhere in the search space.

If, for a given solution, we wish to discover whether a better one may be found elsewhere, it is necessary for the search process to be able to move `down hill' from the local peak, in the hope that, after some backtracking, a route may be found to something which is `higher' than the local peak. Alternatively, search may proceed along two or more parallel or pseudoparallel paths and should be allowed to continue even when one or more peaks have been found - to allow for the possibility that paths which by-pass the peaks which have been found may lead to other peaks which are even higher.

In the present context, the need to be able to escape from local peaks means that the search process needs to be able to create alignments yielding relatively poor compression of New in terms of Old. It has been found in practice with the SP61 and earlier models (see Section \ref{WORKING-MODEL}, below) that, notwithstanding the conjecture on which this research is based (presented at the beginning of Section \ref{SP-THEORY}), it can be advantageous on occasions for the system to create alignments in which IC is very small or even negative.

\subsection{`New' and `Old' and established concepts in computing}

In case concepts like `New' and `Old' seem too far removed from computing as normally understood, it may be useful at this stage to indicate the direction in which these proposals are going. To this end, Table \ref{NEW-OLD} shows how, in the current proposals, concepts of New and Old may be mapped on to established concepts in computing.

\begin{table}
\centering
\begin{tabular}{l l l}
\it Area of & \it New &	\it Old \\
\it application \\
\\
Unsupervised &	`Raw' data. &		Grammar or other \\
inductive learning \space \space & &	knowledge structure \\
 & &					created by learning. \\
\\
Parsing	&	The sentence  &		The grammar used for \\
 &		to be parsed. &		parsing. \\
\\
Pattern & 	A pattern to be &	The stored knowledge \\
recognition &	recognised & 		used to recognise \\
and scene &	or scene to be &	one pattern or several \\
analysis &	analysed. &		within a scene. \\
\\
Databases &	A `query' in SQL or &	Records stored \\	
 &		other query language. \space \space & in the database. \\
\\
Expert & 	A `query' in the &	The `rules' or other \\
system &	query language for &	knowledge stored in \\
 &		the expert system. &	the expert system. \\
\\
Computer & 	The `data' or & 	The computer program \\
program &	`parameters' &		itself. \\
 &		supplied to the \\
 &		program on each run. \\
\end{tabular}

\caption{\small The way in which the concepts of `New' and `Old' in this research appear to relate to established concepts in computing.}
\label{NEW-OLD}
\normalsize
\end{table}

\subsubsection{Unsupervised inductive learning}

In keeping with remarks at the beginning of Section \ref{SP-THEORY} about the origins of the MLE principle as a key to understanding unsupervised inductive learning, this kind of learning - shown first in Table \ref{NEW-OLD} - provides the overall framework for these ideas.

It is envisaged that compression of any reasonably large body of information will normally be done in an incremental manner, processing one section at a time:

\begin{itemize}
\item In keeping with the use of heuristic techniques to accommodate very large search spaces, incremental processing means that, at any one time, the space of possible alignments can be significantly smaller than if the body of information is processed all at once.

\item This incremental approach to compression is convenient also when New information is supplied in an incremental manner as it often is. In this case, processing can start immediately without the need to wait until all the information has been received.
\end{itemize}

In this general framework, it is envisaged that, initially, Old is empty. The first section of New is received or selected and immediately added to Old. It is then processed and compressed as far as possible in terms of patterns in Old. At this early stage, the only compression that can be achieved is due to redundancies that may be present within the first section from New.

In all subsequent stages, processing is done in the same way: each new section of New is added to Old, a set of multiple alignments are formed between New and Old and the best one is selected as a basis for encoding and compressing New in terms of Old. As successive sections of New are processed, the range of patterns in Old will gradually increase so that, generally, it will become progressively easier to find good matches and useful unifications.

At every stage, the system attempts to encode New information using `codes' or `tags' which are attached to patterns in Old. If suitable tags are not present, the system creates new ones and attaches them as required to patterns or parts of patterns in Old.

\subsubsection{Other aspects of computing}

It is envisaged that each of the other manifestations of computing which are shown in Table \ref{NEW-OLD}, and others, may be understood as one {\it part} of the inductive learning process described in the last subsection. Each of these kinds of computing may be seen as being analogous to the processing of one section of New so that, via the formation of MAs, it is compressed in terms of existing patterns in Old using existing codes. The example in Section \ref{SIMPLE-EXAMPLE} illustrates this idea.

\subsection{A working model}\label{WORKING-MODEL}

With the exclusion of the processes for laying down new patterns in Old, modifying existing patterns in Old and attaching new codes in appropriate places, the concepts which have been outlined above are embodied in a software model, SP61, developed as a test-bed for these ideas and as a vehicle for demonstrations. It is envisaged that the next phase in the development of the model will give it a capability for unsupervised inductive learning: the ability to lay down new patterns in Old, modify existing patterns in Old and assign new codes in appropriate places.

A description of the model in its current form is presented in \cite{r22} with a somewhat less detailed description in \cite{r21a}. A fuller description of a similar but slightly earlier version of the model may be found in \cite{r25a} with a briefer description in \cite{r25}. Examples from these and earlier versions may be found in the sources cited in Section \ref{SCOPE}. All the examples of alignments which are shown later in this article are output from SP61, edited as necessary to fit each alignment onto the page.

When the concepts and the model are more mature, it is intended that the model will form the basis for the development of an SP system for practical applications.

\subsection{Related ideas}

Since the primary goal in the development of the ICMAUS framework has been to integrate concepts across a wide range of subfields of computing, it is not altogether surprising that the framework has a family resemblance to other systems. The ability of the ICMAUS framework to model a range of kinds of computing systems was detailed in Section \ref{SCOPE}.

This section notes of some systems with points of similarity with the ICMAUS scheme:

\begin{itemize}
\item The GAMMA parallel programming model \cite{r9a} is based on a chemical reaction model in which the datastructure is a bag or multiset. The computations are interpreted as a succession of chemical reactions consuming the elements of the multiset and producing new elements according to specific rules. Computations are achieved via two basic mechanisms: searching for elements of the multiset that satisfy the reaction conditions and, for each element found, the execution of relevant actions. Any number of reactions may be performed in parallel provided they operate on disjoints sets of elements. Where the reaction conditions select two or more elements, the choice of elements may be made probabilistically.

\item A `classifier system' \cite{r6a} comprises a set of production rules operating in a high-parallel environment, a `bucket brigade' learning algorithm for adjusting weights on rules and a genetic algorithm for the learning of new rules.

\item More generally, rule-based programming is a simple scheme with a wide range of applications (See, for example, \cite{r7a}). Computations are achieved by the interaction of relatively simple elements (rules) in an `object space' that produce new elements or the same elements with modified attributes. Within disjoint subsets of the object space, interactions may proceed in parallel. Pattern matching is the basis of how elements are selected. As with the GAMMA model, rule-based programming may be adapted to operate in a non-deterministic, probabilistic manner.

\item The TREAT algorithm \cite{r10a} is a matching algorithm designed for
efficient pattern matching in a production rule system, running on a
specific high-parallel architecture (`DADO'). As with other systems described above,
pattern matching and a high-parallel environment are prominent features of the system.
\end{itemize}

The ICMAUS framework resembles these systems in the use of many relatively simple standardised entities as the basis of knowledge representation and in the emphasis on pattern matching in the way knowledge is processed. The high-parallel environment finds an echo in the high-parallel form of the projected `SP' computing system, based on the ICMAUS abstract framework. The probabilistic dimension of the ICMAUS scheme is another feature shared with schemes like those just mentioned.

Perhaps the most important difference between these systems and the ICMAUS framework is in the intended scope of the system. By conjecture, the scope of the ICMAUS framework is as broad as computing itself: nothing is taken for granted including such seamingly basic notions as `variable', the number system, negation, comparisons like `greater than' or `less than', and so on. The research aims, amongst other things, to provide an interpretation of concepts like these. Also distinctive of the ICMAUS framework is the emphasis on IC as a unifying theme and, by conjecture, unification of matching patterns as the sole mechanism for achieving IC. Other differences are that `patterns' in the ICMAUS scheme are even simpler than production rules (patterns lack a rewrite arrow and are without any distinction between `condition' and `action') and that multiple alignment is prominent in the ICMAUS framework in a way that it is not in other systems. More controversially, perhaps, the probabilistic dimension of the ICMAUS scheme \cite{r21a, r22, r23, r24} is intrinsic to the basic conceptual framework in a more fundamental way, perhaps, than it is in other schemes.

\subsubsection{DNA computing}\label{DNA-COMPUTING}

\cite{r0a} demonstrated that simple versions of the Hamiltonian path problem could be solved using DNA in solution (see also \cite{r0}). Given the prominence of multiple alignment in the ICMAUS framework and given that MA is often associated with the analysis of DNA molecules, one might naturally assume a relationship between DNA computing and the ICMAUS scheme.

The main significance of DNA computing in relation to the ICMAUS concepts is probably that chemical reactions with DNA (and other molecules) provide a means of matching patterns and this may be done with massive parallelism. Thus potentially, this mode of computing provides a vehicle for the pattern matching required in the ICMAUS scheme and the high levels of parallelism which may help to achieve useful speeds.

In Adleman's first experiments the pattern matching was an all-or-nothing kind, not in itself appropriate for the kind of partial matching required for ICMAUS in its more fully developed forms. However, later work \cite{r2a} has shown that pattern matching with DNA can support the more sophisticated kinds of heuristic processing needed to find good partial matches between patterns. 

\section{Universal Turing Machine and Post Canonical System}\label{UTM-PCS}

In its standard form \cite{r19}, a UTM comprises a `tape' on which symbols may appear in sequence, a `read-write head' and a `transition function'. In each of a sequence of steps, a symbol is read from the tape at the current position of the head and then, depending on that symbol, on the transition function and the current `state' of the head, a new symbol is written on the tape, the state of the head is changed, and the head moves one position left or right.

A more detailed description of UTMs is not needed here because the arguments to be presented take advantage of the fact \cite[chapters 10 to 14]{r10} that the operation of any UTM can be modelled with a PCS \cite{r11}. In this section, I shall describe the workings of a PCS and then, in Section \ref{ICMAUS-AND-PCS}, I shall try to show how the operation of a PCS, and thus a UTM, may be understood in terms of ICMAUS.

\subsection{The structure of a PCS}\label{STRUCTURE-PCS}

A PCS comprises:

\begin{itemize}
\item An {\it \bf alphabet} of primitive {\it \bf symbols} (`letters' in Post's terminology),

\item One or more {\it \bf primitive assertions} or {\it \bf axioms} (strings of symbols which can often be regarded as `input').

\item One or more {\it \bf productions} which can often be regarded as a `program'.

\item Although this is not normally included in formal definitions of the PCS concept, it is clear from how the system is intended to work (see Section \ref{HOW-PCS-WORKS} below), that the PCS also includes mechanisms for {\it \bf matching patterns} and for {\it \bf searching} for matches between leading substrings in the input pattern and the left-hand ends of the productions.
\end{itemize}

Each production has this general form:\footnote{Spaces between symbols here and in other examples have been inserted for the sake of readability and because it allows us to use atomic symbols where each one comprises a string of two or more characters (with spaces showing the start and finish of each string).
Otherwise, spaces may be ignored.}

\[g_0\ \$_1\ g_1\ \$_2\ ...\ \$_n\ g_n \rightarrow  h_0\ \$^{\prime}_1\ h_1\ \$^{\prime}_2\ ...\ \$^{\prime}_m\ h_m\]

\noindent where ``Each $g_i$ and $h_i$ is a certain fixed string; $g_0$ and $g_n$ are often null, and some of the $h$'s can be null. Each $\$_i$ is an `arbitrary' or `variable' string, which can be null. Each $\$^{\prime}_i$ is to be replaced by a certain one of the $\$_i$.'' \cite[pp. 230-231]{r10}.

In its simplest `normal' form, a PCS has one primitive assertion and each production has the form:

\[g\ \$ \rightarrow  \$\ h\]

\noindent where $g$ and $h$ each represent a string of zero or more symbols, and both instances of `\$' represent a single `variable' which may have a `value' comprising a string of zero or more symbols.

It has been proved \cite{r11} that any kind of PCS can be reduced to a PCS in normal form (see also \cite[chapter 13]{r10}). That being so, this kind of PCS will be the main (but not exclusive) focus of our attention.

\subsection{How the PCS works}\label{HOW-PCS-WORKS}

When a PCS (in normal form) processes an `input' string, the first step is to find a match between that string and the left-hand side of one of the productions in the given set. The input string matches the left hand side of a production if a match can be found between leading symbols of the input string and the fixed string (if any) at the start of that left-hand side, with the assignment of any trailing substring within the input string to the variable within the left-hand side of the production.

Consider, for example, a PCS comprising the alphabet `a ... z', an axiom or input string `a b c b t', and productions in normal form like this:

\begin{center}
a \$ $\rightarrow$ \$ a \linebreak
b \$ $\rightarrow$ \$ b \linebreak
c \$ $\rightarrow$ \$ c \linebreak
\end{center}

In this example, the first symbol of the input string matches the first symbol in the first production, while the trailing `b c b t' is considered to match the variable which takes that string as its value. The result of a successful match like this is that a new string is created in accordance with the configuration on the right hand side of the production which has been matched. In the example, the new string would have the form `b c b t a', derived from `b c b t' in the variable and `a' which follows the variable on the right hand side of the production.

After the first step, the new string is treated as new input which is processed in exactly the same way as before. In this example, the first symbol of `b c d t a' matches the first symbol of the second production, the variable in that production takes `c d t a' as its value and the result is the string `c b t a b'.

This cycle is repeated until matching fails. It should be clear from this example that the effect would be to `rotate' the original string until it has the form `t a b c b'. The `t' which was at the end of the string when processing started has been brought round to the front of the string. This is an example of the ``rotation trick'' used by \cite[Chapter 13]{r10} in demonstrating how a PCS in normal form can model any kind of PCS.

With some combinations of alphabet, input and productions, the process of matching strings to productions never terminates. With some combinations of alphabet, input and productions, the system may follow two or more `paths' to two or more different `conclusions' or may reach a given conclusion by two or more different routes. The `output' of the computation is the set of strings created as the process proceeds.

\subsection{Other examples}\label{OTHER-EXAMPLES}

For readers unfamiliar with this kind of system, two other examples are included here to show more clearly how a PCS may achieve `computing'. The first example is based on two examples from \cite[Chapter 12]{r10} showing how odd and even unary numbers may be generated. The second example is from the same source. Other examples may be found in the same chapter.

\subsubsection{Generating numbers in unary notation}\label{GENERATING-UNARY-NOS}

In the unary number system, 1 = `1', 2 = `1 1', 3 = `1 1 1' and so on. The recursive nature of the unary number system can be expressed very clearly and succinctly with a PCS like this:

\begin{itemize}
\item Alphabet: the single symbol 1.
\item Axiom: 1.
\item Production: If any string `\$' is a number, then so is the string `\$ 1'. 
This can be expressed with the production:
\end{itemize}

\[\$ \rightarrow \$\ 1\]

It should be clear from the description of how the PCS works, above, that this PCS will generate the infinite series of unary strings: `1', `1 1', `1 1 1', `1 1 1 1', `1 1 1 1 1' etc. 

Less obvious, perhaps, is that the same PCS can be used to recognise a string of symbols as being an example of a unary number. This is done by using the production in `reverse', matching a character string to the right hand side of the production, taking the left hand side as the `output' and then repeating the right-to-left process until only the axiom will match.

\subsubsection{Generating palindromes}\label{GENERATING-PALINDROMES}

To quote \cite[p. 228]{r10}: ``The `palindromes' are the strings that read the same backwards and forwards, like {\it cabac} or {\it abcbbcba}. Clearly, if we have a string that is already a palindrome, it will remain so if we add the same letter to the beginning and end. Also, clearly, we can obtain all palindromes by building in this way out from the middle.'' Here is his example:

\begin{itemize}
\item Alphabet: a, b, c.
\item Axioms: a, b, c, aa, bb, cc.
\item Productions:
\end{itemize}

\begin{center}
\$ $\rightarrow$ a \$ a \linebreak
\$ $\rightarrow$ b \$ b \linebreak
\$ $\rightarrow$ c \$ b \linebreak
\end{center}

Although this example is not in normal form, it should be clear how it generates all the palindromes for the given alphabet. As with the previous example, this PCS may be used to recognise whether or not a string of characters is a palindrome by using the productions in `reverse'.

\section{ICMAUS and the operation of a PCS}\label{ICMAUS-AND-PCS}

It is not hard to see that a successful match between a leading substring in an input string and the leading substring in a PCS production may be interpreted in terms of the alignment and unification of patterns, as outlined in Section \ref{SP-THEORY}. What is less obvious is how concepts of alignment and unification might be applied to the processes of matching a trailing string of symbols to the variable, assigning it as a value to the variable and incorporating it in an output string.

\subsection{Modelling one step of the ``rotation trick''}\label{AN-EXAMPLE}

In the proposed SP framework (ICMAUS), the first PCS described above (for the `rotation' of a string) may be modelled using the patterns shown in Figure \ref{ROTATE-PATTERNS}. Regarding these patterns:

\begin{itemize}
\item The pattern under the heading `New' corresponds to the axiom or `input'.

\item The four patterns on the left under the heading `Old' may be seen as a representation of the alphabet. 

\item The first three patterns on the right correspond to the three productions of the PCS. As we shall see, the pair of symbols, `\$ \#\$', in each pattern function as if they were a `variable'. There is no rewrite arrow (`$\rightarrow$') and only one instance of the `variable'.

\item As we shall see, the last pattern on the right (`\$ L \#L \$ \#\$ \#\$') is, in effect, a recursive definition of the concept of `any string of the letters in the given alphabet'. This is comparable with the production shown in the example in Section \ref{GENERATING-UNARY-NOS} which expresses the recursive nature of the unary number system. It is, in effect, a `type definition' for the variable defining the infinite set of alternative strings (composed from the given alphabet) which the variable may take.\footnote{Readers will note that the pattern
`\$ L \#L \$ \#\$ \#\$' contains two `variables': the pair of symbols `\$ \#\$', already considered, and also the pair of symbols `L \#L' which can take any of `a', `b' or `c' as its `value'.}
\end{itemize}

In these patterns, none of the symbols have any `hidden' meaning. In particular, the symbols `\$', `\#\$', `L', `\#L', `P' and `\#P' (which may be described, informally, as `service' symbols) have exactly the same formal status as other symbols (`a', `b' etc) and enter into matching and unification in the same way.\footnote{Although it is true in general for the ICMAUS framework that all symbols have the same status, the principle should perhaps be qualified here since, at the end of each cycle, it is proposed below that the `service' symbols are removed from the `output' string leaving only the `data' symbols for further processing.}

\begin{figure}
\centering
\begin{tabular}{l l}
{\it New} \\
\\
a b c b t \\
\\
{\it Old} \\
\\
L\ a\ \#L\ \ \ \ \ \ \ & P\ a\ \$\ \#\$\ a\ \#P \\
L\ b\ \#L\ \ \ \ \ \ \ & P\ b\ \$\ \#\$\ b\ \#P \\
L\ c\ \#L\ \ \ \ \ \ \ & P\ c\ \$\ \#\$\ c\ \#P \\
L\ t\ \#L\ \ \ \ \ \ \ & \$\ L\ \#L\ \$\ \#\$\ \#\$ \\
\end{tabular}

\caption{\small SP patterns modelling the first example of a PCS described in Section \ref{STRUCTURE-PCS}. An interpretation of these patterns is given in the text.}
\label{ROTATE-PATTERNS}
\normalsize
\end{figure}

\subsubsection{One step}

\begin{figure}
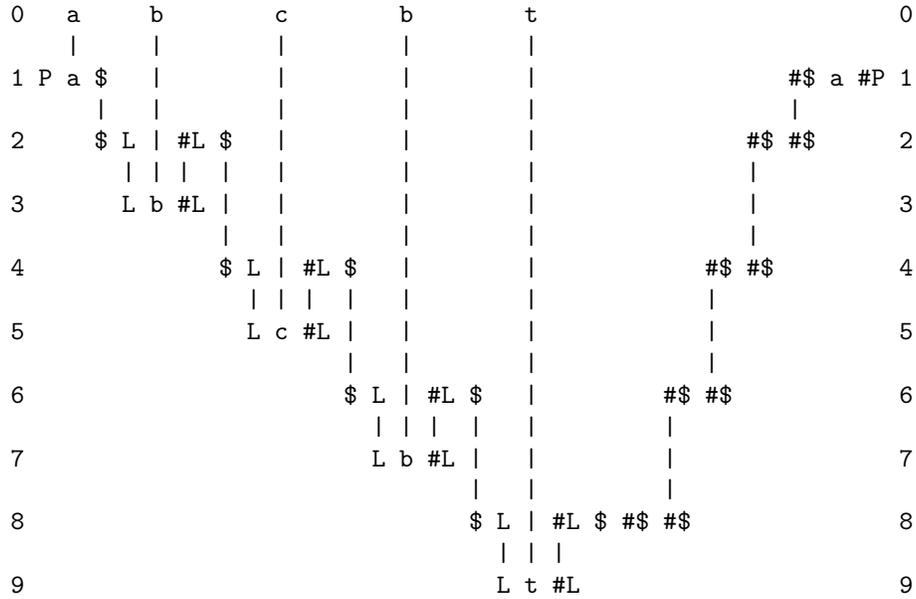

\centering
\begin{BVerbatim}
0   a     b        c        b        t                          0
    |     |        |        |        |                           
1 P a $   |        |        |        |                  #$ a #P 1
      |   |        |        |        |                  |        
2     $ L | #L $   |        |        |               #$ #$      2
        | | |  |   |        |        |               |           
3       L b #L |   |        |        |               |          3
               |   |        |        |               |           
4              $ L | #L $   |        |            #$ #$         4
                 | | |  |   |        |            |              
5                L c #L |   |        |            |             5
                        |   |        |            |              
6                       $ L | #L $   |         #$ #$            6
                          | | |  |   |         |                 
7                         L b #L |   |         |                7
                                 |   |         |                 
8                                $ L | #L $ #$ #$               8
                                   | | |                         
9                                  L t #L                       9
\end{BVerbatim}
\caption{\small The best alignment (in terms of IC) found by SP61 using the patterns shown in Figure \ref{ROTATE-PATTERNS} as described in the text.}
\label{ROTATE-ALIGNMENT}
\normalsize
\end{figure}

Figure \ref{ROTATE-ALIGNMENT} shows the best alignment (in terms of IC) found by SP61 with the `input' pattern in New and the other patterns in Old. In the figure, the repeated appearances of the pattern `\$ L \#L \$ \#\$ \#\$' represent a single instance of that pattern in accordance with the generalisation of the multiple alignment problem described in Section \ref{GENERALISATION-MA}. The same is true of the two appearances of the pattern `L b \#L'.

Unification of the matching symbols in Figure \ref{ROTATE-ALIGNMENT} has the effect of `projecting' the alignment into a single sequence, thus:

\begin{center}
P a \$ L b \#L \$ L c \#L \$ L b \#L \$ L t \#L \$ \#\$ \#\$ \#\$ \#\$ \#\$ a \#P.
\end{center}

\noindent Notwithstanding the interpolation of instances of `service' symbols like `P', `\$', `L', `\#L' etc, this sequence contains the subsequence `a b c b t' corresponding to the `input' and also the subsequence `b c b t a' corresponding to the `output' of the first step of our first example of a PCS.

In this example, the subsequence `b c b t' corresponding to the contents of the `variable' is shared by the input and the output thus removing the need for the rewrite arrow in the PCS. An analogy for this idea is the way Welsh and English are sometimes conflated in bilingual signs in Wales. The phrase `Cyngor Gwynedd' in Welsh means `Gwynedd Council' in English. So it is natural, in bilingual signs put up by Gwynedd Council, to represent both the Welsh and the English forms of the name using the words `Cyngor Gwynedd Council'. In this example, the word `Gwynedd' serves both the Welsh version of the name and the English version in much the same manner that, in our example above, the contents of the variable serves both the input string and the output string.

\subsubsection{Repetition of steps}

The second and subsequent steps may proceed in the same way. At the end of each step, we may suppose that the `service' symbols are stripped out and the sequence from the contents of the `variable' onwards is presented to the system again as New. When `t' reaches the front of the string, there is no production with a leading `t' which means that New at that stage cannot be fully matched. In this example, this may be taken as a signal that the rotation is complete.

In Minsky's example (which takes too much space to reproduce here), the presence at the beginning of the string of `T1' (which may be equated with `t' in our example) causes other productions to `fire', leading the processing into other paths.

\subsection{Other examples}

This subsection shows how the second and third examples of a PCS which were described above may be modelled as ICMAUS.

\subsubsection{Unary numbers and ICMAUS}

Figure \ref{UNARY-PATTERNS} shows two patterns to model the example of a PCS for generating unary numbers (Section \ref{GENERATING-UNARY-NOS}, above). The first pattern corresponds with the axiom in the PCS and the second one corresponds with the production in the PCS.

\begin{figure}
\centering
\begin{tabular}{l}
{\it New} \\
\\
1 (Other options for New are described in the text) \\
\\
{\it Old} \\
\\
\$\ \$ \#\$\ 1\ \#\$ \\
\end{tabular}
\caption{\small Patterns for processing by SP61 to model a PCS to produce or recognise unary numbers (Section \ref{GENERATING-UNARY-NOS}).}
\label{UNARY-PATTERNS}
\normalsize
\end{figure}

If the two symbols `\$ \#\$' in the body of the pattern `\$ \$ \#\$ 1 \#\$' have nothing between them, the pattern represents `1' by itself. Otherwise, the first and last symbols in the pattern can be matched with the same pair in the body of the pattern - and this pair is followed by `1'. Thus the pattern `\$ \$ \#\$ 1 \#\$' may be read as ``A unary number is `1' or any unary number followed by `1'''.

If the first of the two patterns is supplied to SP61 as New and the second is placed in Old, the program will generate a succession of `good' alignments, one example of which is shown in Figure \ref{UNARY-ALIGNMENTS} (a). If it is not stopped, the program will continue producing alignments like this until the memory of the machine is exhausted.

If we project the alignment in Figure \ref{UNARY-ALIGNMENTS} (a) into a single sequence and then ignore the `service' symbols in this example (`\$' and `\#\$'), we can see that the system has, in effect, generated the unary number `1 1 1 1 1'. We can see from this example how the alignment has captured the recursive nature of the unary number definition.

Readers may wonder whether it makes any sense to `compress' something as small as `1' using a relatively large pattern like `\$ \$ \#\$ 1 \#\$' which itself contains `1', especially if it is repeated several times in an alignment. Relevant discussion may be found in Section \ref{OTHER-EXAMPLES}, below.

\begin{figure}
\centering
\begin{BVerbatim}
0                1                        0
                 |                         
1         $ $ #$ 1 #$                     1
          |        |                       
2       $ $        #$ 1 #$                2
        |               |                  
3     $ $               #$ 1 #$           3
      |                      |             
4   $ $                      #$ 1 #$      4
    |                             |        
5 $ $                             #$ 1 #$ 5

(a)

0                1    1    1    1    1    0
                 |    |    |    |    |     
1         $ $ #$ 1 #$ |    |    |    |    1
          |        |  |    |    |    |     
2       $ $        #$ 1 #$ |    |    |    2
        |               |  |    |    |     
3     $ $               #$ 1 #$ |    |    3
      |                      |  |    |     
4   $ $                      #$ 1 #$ |    4
    |                             |  |     
5 $ $                             #$ 1 #$ 5

(b)
\end{BVerbatim}
\caption{\small (a) One of many `good' alignments produced by SP61 from the patterns in Figure \ref{UNARY-PATTERNS}. (b) The best alignment (in terms of compression) produced by SP61 with `1 1 1 1 1' in New and, in Old, the patterns from `Old' in Figure \ref{UNARY-PATTERNS}.}
\label{UNARY-ALIGNMENTS}
\normalsize
\end{figure}

Figure \ref{UNARY-ALIGNMENTS} (b) shows the best alignment produced by SP61 when `1' in New is replaced by `1 1 1 1 1'. This alignment is, in effect, a recognition of the fact that `1 1 1 1 1' is a unary number. It corresponds to the way a PCS may be run `backwards' to recognise input patterns but, since there is no left-to-right arrow in the ICMAUS scheme, the notion of `backwards' processing does not apply. Other unary numbers may be recognised in a similar way.

\subsubsection{Palindromes and ICMAUS}

Figure \ref{PALINDROME-PATTERNS} shows patterns that may be used with SP61 to model the PCS described in Section \ref{GENERATING-PALINDROMES}. The first six patterns in Old may be seen as analogues of the six axioms in the PCS. The last pattern in Old expresses the recursive nature of palindromes. The pattern above it serves to link the single letter patterns (the first three in Old) to the recursive pattern.

\begin{figure}
\centering
\begin{tabular}{l}
{\it New} \\
\\
a c b a b a b c a \\
\\
(Other options for New are described in the text) \\
\\
{\it Old} \\
\\
L a \#L \\
L b \#L \\
L c \#L \\
L1 a \#L1 L2 a \#L2 \\
L1 b \#L1 L2 b \#L2 \\
L1 c \#L1 L2 c \#L2 \\
\$ L \#L \#\$ \\
\$ L1 \#L1 \$ \#\$ L2 \#L2 \#\$ \\
\end{tabular}
\caption{\small Patterns for processing by SP61 to model a PCS to produce or recognise palindromes (Section \ref{GENERATING-PALINDROMES}).}
\label{PALINDROME-PATTERNS}
\normalsize
\end{figure}

\begin{figure}
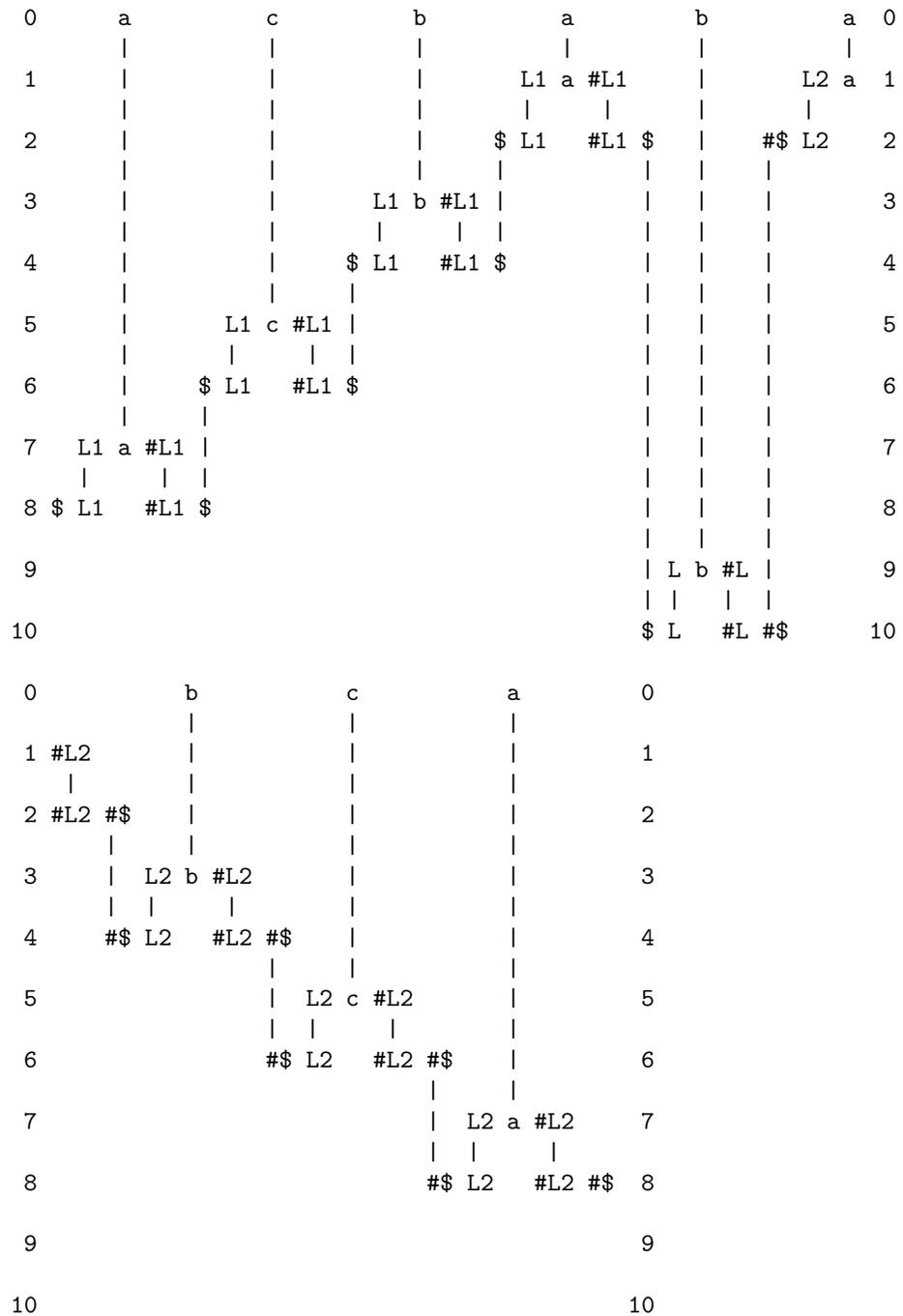

\centering
\begin{BVerbatim}
 0      a          c          b          a         b          a  0
        |          |          |          |         |          |   
 1      |          |          |       L1 a #L1     |       L2 a  1
        |          |          |       |     |      |       |      
 2      |          |          |     $ L1   #L1 $   |    #$ L2    2
        |          |          |     |          |   |    |         
 3      |          |       L1 b #L1 |          |   |    |        3
        |          |       |     |  |          |   |    |         
 4      |          |     $ L1   #L1 $          |   |    |        4
        |          |     |                     |   |    |         
 5      |       L1 c #L1 |                     |   |    |        5
        |       |     |  |                     |   |    |         
 6      |     $ L1   #L1 $                     |   |    |        6
        |     |                                |   |    |         
 7   L1 a #L1 |                                |   |    |        7
     |     |  |                                |   |    |         
 8 $ L1   #L1 $                                |   |    |        8
                                               |   |    |         
 9                                             | L b #L |        9
                                               | |   |  |         
10                                             $ L   #L #$      10

 0           b           c           a         0
             |           |           |          
 1 #L2       |           |           |         1
    |        |           |           |          
 2 #L2 #$    |           |           |         2
       |     |           |           |          
 3     |  L2 b #L2       |           |         3
       |  |     |        |           |          
 4     #$ L2   #L2 #$    |           |         4
                   |     |           |          
 5                 |  L2 c #L2       |         5
                   |  |     |        |          
 6                 #$ L2   #L2 #$    |         6
                               |     |          
 7                             |  L2 a #L2     7
                               |  |     |       
 8                             #$ L2   #L2 #$  8
                                                
 9                                             9
                                                
10                                            10
\end{BVerbatim}
\caption{\small \sloppy The best alignment (in terms of compression) produced by SP61 with the patterns from Figure \ref{PALINDROME-PATTERNS} in New and Old, as shown in that figure.}
\label{PALINDROME-ALIGNMENT}
\normalsize
\end{figure}

Figure \ref{PALINDROME-ALIGNMENT} shows the best alignment produced by SP61 with `a c b a b a b c a' in New and, in Old, the patterns under the heading `Old' in Figure \ref{PALINDROME-PATTERNS}. The alignment may be seen as a recognition that the pattern in New is indeed a palindrome. Other palindromes (using the same alphabet) may be recognised in a similar way.

As with the example of unary numbers, the same patterns from Old may be used to generate palindromes. In this case, some kind of `seed' is required in New like a single letter (`a', `b' or `c') or a pair of letters (`a a', `b b' or `c c'). Given one of these `seeds' in New, the system can generate palindromes until the memory of the machine is exhausted.

Readers may wonder whether the patterns from Old in Figure \ref{PALINDROME-PATTERNS} might be simplified. Would it not be possible to substitute `\$ a \#\$' for `L a \#L' (and likewise for the other single letter `axioms') and to replace `L1 a \#L1 L2 a \#L2' with `\$ a \$ \#\$ a \#\$' (and likewise for the other pairs) and then remove `\$ L \#L \#\$' and `\$ L1 \#L1 \$ \#\$ L2 \#L2 \#\$'? This is possible but the penalty is that, for reasons that would take too much space to explain, searching for the best alignments takes very much longer and leads along many more `false trails'.

\subsection{Information compression and the operation of a PCS}

It should be clear from the foregoing discussion and examples that the process of matching a string of symbols with one production in a PCS may always be modelled as the formation of an alignment as defined in this research.

Given the conjecture which provides the inspiration for this research (described at the beginning of Section \ref{SP-THEORY}), a pertinent question to ask is: ``Does every alignment of this kind always yield compression of New in terms of patterns in Old?''

The answer to this question depends on the details of how the encoding of New in terms of Old is done, including assumptions on which any given method is based. It is certainly possible to devise encoding methods where the answer to the question, above, would be ``no''. With the method used in SP61 (outlined in Section \ref{EVALUATION-OF-ALIGNMENTS}), the answer is ``yes'', provided that a key assumption is made, described and discussed in the following subsections.

\subsubsection{When an encoding has fewer symbols than New}

Let us consider, first of all, an example where the conclusion is likely to be uncontroversial even with very conservative assumptions. Let us suppose that New contains the sequence `a l o n g s e q u e n c e o f s y m b o l s' and that Old contains, amongst others, the pattern `P 1 a l o n g s e q u e n c e o f s y m b o l s \#P'.

With this example, it should be clear that New can be aligned with the matching sequence of symbols in the pattern from Old and that New may be encoded in terms of this pattern as something like `P 1 \#P'. The encoding is much smaller than New in terms of numbers of symbols and, if we assume that a fixed number of bits is used for the encoding of every symbol, there is also compression in terms of bits of information.

\subsubsection{When an encoding has as many or more symbols than New}

If New contains only one symbol, `a' and if Old contains patterns like `P 1 a \#P', `P 2 b \#P', and so on, one would naturally assume that no positive compression could be achieved. If `a' in New is aligned with `a' in `P 1 a \#P', then its encoding would be `P 1 \#P' or something similar. If a fixed number of bits is used for the encoding of every symbol, then each of these encodings represents negative compression.

In SP61, it is assumed that every symbol in New is larger than the theoretical minimum needed to discriminate one symbol from another - which means, in effect, that each symbol in New is assumed to contains redundancy. This is like assuming that each symbol in New represents a relatively large chunk of information like `j o h n' in the example in Section \ref{SIMPLE-EXAMPLE}. The assumption that each symbol in New is relatively large is coupled with a calculation of two sizes for symbols in New: a notional `actual' size which is relatively large and a `minimum' size which is the size of a code which may be used to represent the symbol and is close to the theoretical minimum. For clarity in this discussion, each symbol from New in its actual size will be represented directly as `a', `b' etc, while the code for a symbol from New will be represented with the same characters and `prime', e.g., `a$^{\prime}$', `b$^{\prime}$' etc.

Let us suppose that `a b c b t' in New is aligned with patterns from Old as shown in Figure \ref{ROTATE-ALIGNMENT}. In this case, the encoding of New in terms of Old would be `P a$^{\prime}$ b$^{\prime}$ c$^{\prime}$ b$^{\prime}$ t$^{\prime}$ \#P' or something similar. This encoding has more symbols than the sequence from New which is being encoded so, on a superficial reading, there should be zero or negative compression. But if the `actual' size of each symbol from New is, say, 20, bits, if the `minimum' size of each symbol from New and each of `P' and `\#P' is, say, 4 bits, then the encoding above represents a compression of $(5 \times 20) - (7 \times 4) = 72$ bits.

\subsection{Redundancy in the operation of digital systems}

The assumption, above, that each symbol in New contains redundancy may seem like an {\it ad hoc} assumption to salvage the conjecture on which this research is based. However, a little reflection may convince us that the assumption has some justification.

\subsubsection{Computing with `0' and `1'}

One of the attractions of digital systems is that they are relatively robust in the face of noise. If, as is normally the case, a binary alphabet is used, the two symbols (normally `0' and `1') are typically represented by two different voltages whose values are different enough to ensure that the electronics can discriminate them with little or no error through billions of computations.

Given that the technology is capable of discriminating a much larger number of different voltages than two, the use of only two voltages is analogous to the use of only `A' and `Z' from an alphabet of 26 symbols. Since an alphabet of 26 symbols requires, in round figures, $\log_{2}26 \bumpeq 5$ bits of information for each symbol (assuming that they are equiprobable), the two symbols `A' and `Z' are expressed with 5 times as many bits as the theoretical minimum which is 1.

If the analogy is accepted as valid, then we may conclude that digital computing is based on the provision of redundancy in the `bedrock' of its computations. Reducing the redundancy at this level (by bringing the difference in voltages closer to the threshold of what the electronics can discriminate) would cause errors to increase. Would any computing be possible if all redundancy at this level were eliminated?

\subsubsection{Identifiers, addresses and codes}

If we consider `higher' levels in the organisation of digital computers, we find that there is typically much redundancy in how entities of different kinds are identified.

This is clearly true of identifiers like the names of files, names of tables and fields in databases, and the names of functions, variables and so on in computer programs. It may be objected with some justice that this redundancy exists primarily to serve the psychological needs of humans who must read and remember these identifiers and that the identifiers are all converted into addresses on disk or in computer memory which is where `real' computing is done.

But addresses to locations in computer memory also contain redundancy. It is rare for any application to use every location which has been provided, so more bits are needed to identify each location than are strictly necessary. It may be objected, with justice, that this is because, with current technology, we cannot provide new memory locations only when required. More to the point, perhaps, is that some memory locations are likely to be used more often than others but the addressing system does not take advantage of this by assigning short addresses to frequently-accessed locations and longer addresses to locations which are accessed rarely. Indirect addressing may go partly down this road but it is only an approximate solution.

In a similar way, fixed-length codes for symbols or other entities (e.g., the 8-bit codes of the ASCII character set) are convenient but are normally quite redundant because they do not exploit variations in the frequencies with which different codes are used.

\subsubsection{Huffman coding}

Of course, there are well-known techniques for constructing variable-length codes to take advantage of variations in frequency. The best-known of these is Huffman coding (see \cite{r5}).

It is often said that the sizes of Huffman codes are at the theoretical minimum. In one sense, this is true but it is slightly misleading. In order to `parse' a sequence of binary symbols representing a sequence of Huffman codes, it is necessary to use the trie which was created alongside the Huffman codes. The codes themselves may be minimal but, in themselves, they are unusable. The additional information needed to specify the trie may be seen as redundancy within the whole system.

\subsubsection{Discussion}

This quick review of the workings of digital computers has shown that redundancy pervades the system in the way that symbols and other entities are identified. And even where it seems that all redundancy has been eliminated, it can be argued that it is present in a covert form.

Redundancy equates with `structure'. If a computing system, including its programs, is totally lacking in redundancy then it is totally unstructured - which means that it is totally random. A system which is totally random and unstructured is no system at all. It certainly cannot support the highly structured and predictable kinds of information processing that we know as `computing'. In a similar way, meaningful computing depends on the existence of redundancy at some level in the input data, even if it is only in the symbolic structure of the data.

These considerations suggest that the deliberate introduction of redundancy into the symbols in New in the SP model is more than an {\it ad hoc} `fix' for a theoretical anomaly. It seems that computing depends in a fundamental way on the existence of redundancy at some level both in the structure of the computer and its programs and also in the data to be processed by the computer.

\section{Augmentation of established models of computing}\label{AUGMENTATION}

Section \ref{ICMAUS-AND-PCS} has described, with examples, how the operation of a PCS in normal form may be understood as ICMAUS. Since it is known that any PCS may be modelled by a PCS in normal form (\cite{r11}, \cite[Chapter 13]{r10}), we may conclude, tentatively, that the operation of any PCS may be interpreted in terms of ICMAUS. Since we also know that any UTM may be modelled by a PCS \cite[Chapter 14]{r10} we may also conclude, tentatively, that the operation of any UTM may be interpreted in terms of ICMAUS.

Although the operation of a PCS or UTM may understood in terms of ICMAUS, it does not require the full range of capabilities of a well-developed ICMAUS system. Matching in a PCS or UTM is done in a relatively constrained manner (Section \ref{CONSTRAINTS}) which does not require sophisticated search.

What is envisaged for the proposed SP system is a much more fully developed realisation of ICMAUS mechanisms including well-developed abilities for heuristic search, for finding `good' partial matches between patterns and for building a wide variety of alignments amongst patterns.

Figure \ref{CONVENTIONAL-AND-SP-COMPUTERS} shows a schematic representation of the proposed `SP' computer with a schematic representation of a conventional computer (or UTM or PCS etc) for comparison. The next two subsections consider each of these in turn.

\subsection{Conventional computers (UTM, PCS etc)}

Schematically, the conventional computer (or UTM or PCS etc) comprises the core machine (C), together with software across a range of applications (S).

In terms of the ideas described in Sections 2 and 4, we may suppose that C in the conventional computer may be interpreted as ICMAUS with relatively simple search mechnanisms which operate in a somewhat rigid `clockwork' manner.

In a PCS, searching and matching occurs when the input string is
compared with the left-hand sides of productions. In a UTM, searching
and matching occurs when a symbol newly read from the tape, together with
a symbol representing the state of the read-write head, is compared with
the left-hand part of each of several tuples in the transition
function. Where exact matches are found, the right-hand part(s) of the corresponding tuples control (in a deterministic or non-deterministic manner) what happens next.

In a conventional computer, one manifestation of searching and matching is the process by which an item in memory is accessed using the address of that item. This may be seen as a process of searching through addresses in memory until a match is found for the given address. Accessing an item in memory is served by logic circuits which, with great speed, eliminate areas of memory progressively until the item is found.

This kind of highly constrained form of searching and matching has advantages for certain kinds of problem which can be solved with great accuracy and speed. The penalty is that other kinds of problem which require more flexibility cannot be solved unless additional mechanism is provided in S.

Thus, owing to the limitations of C in the conventional computer, many applications within S contain additional ICMAUS mechanisms designed to make good the deficiencies of the core mechanisms. Similar mechanisms are provided repeatedly in diverse applications. For example, search in various forms is found in database systems, expert systems, parsers, compilers, text editors and DTP software (especially software for spelling checking and correction), spreadsheets, genetic algorithms of various kinds, data mining software, and many other more or less specialised kinds of software. Thus there is much more of this kind of repetition than is suggested by the three instances of ICMAUS shown in the `S' part of the conventional computer as portrayed in Figure \ref{CONVENTIONAL-AND-SP-COMPUTERS}.

As already indicated, ICMAUS mechanisms provided in C of the conventional computer do not exploit the full potential of this type of mechanism. The ICMAUS mechanisms in S are also incomplete because, in general, each one provides only what is required for a specific application and, out of those requirements, they normally provide only what is missing from C. Hence, to show that these instances of ICMAUS mechanisms in C and S of the conventional computer are not fully developed, they are represented in Figure \ref{CONVENTIONAL-AND-SP-COMPUTERS} using relatively small type.

What else is in S? In general, the parts of S which are not ICMAUS mechanisms are domain-specific knowledge about such things as the rules of accounting, the classes of objects found in a particular organisation (each with its associated data structures and methods), the laws of thermodynamics, aerodynamics or topology, and many other concepts across all the many areas where computers can be applied.

\subsection{The proposed `SP' computer}

In the proposed SP computer, the core mechanism (C) comprises ICMAUS mechanisms developed as fully as our understanding and available resources allow. Although C in the SP computer is probably larger and more complex than C in the conventional computer, the anticipated pay-offs are the elimination of repeated instances of ICMAUS in S giving an overall simplification of the system (C + S) and better integration across applications.

The key difference between ICMAUS in C in a conventional computer and ICMAUS in C of an SP computer seems to be that, without additional programming, the former is restricted to exact matches between sequences of symbols whereas the latter should have a well-developed capability for finding good partial matches between sequences of symbols or, in advanced versions, good partial matches between arrays of symbols in two or more dimensions. The core of the projected SP computer should also contain mechanisms for building multiple alignments of the kind described in Section \ref{ICMAUS-AND-PCS}.

\subsubsection{Constraining the SP computer to behave like a conventional computer}

Although the SP computer should be able to find good partial matches between arrays of symbols, it should be possible to restrict it to exact matches between sequences of symbols if that is required for any particular application. Thus, with suitable constraints (see Section \ref{CONSTRAINTS}), the SP computer may be made to function, when required, as if it was a conventional computer.

\begin{figure}
\centering
\includegraphics[width=10cm]{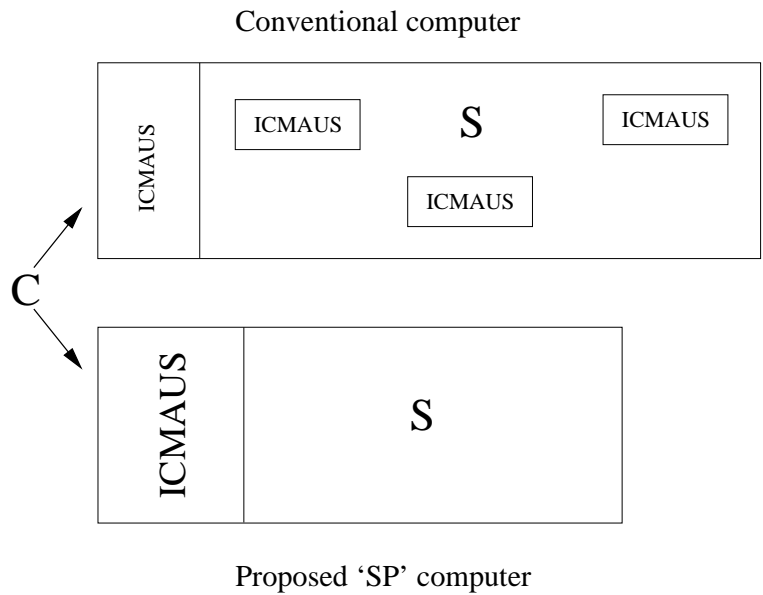}
\caption{\small Schematic representation of a conventional computer and the proposed `SP' computer in terms of the ICMAUS proposals. {\it \bf Key:} C = the `core' of the conventional computer and of the SP computer; ICMAUS (in small type) = ICMAUS mechanisms (described in Section \ref{SP-THEORY}) in various relatively restricted forms. ICMAUS (in large type) = ICMAUS mechanisms in a relatively fully developed and generic form; S = software to cover a wide range of applications. In the conventional computer, S includes many instances of (small) ICMAUS - more than the three instances shown here. It is assumed that the data to be processed is the same for both computers.}
\label{CONVENTIONAL-AND-SP-COMPUTERS}
\end{figure}

\subsection{Minimum Length Encoding in the design of computers?}

The principles just described are similar to the principle of Minimum Length Encoding in grammar induction (described in Section \ref{MLE-LEARNING}).

In the case of grammar induction, the principle requires a search for a grammar which minimises the size of the grammar together with the size of the sample when it is encoded in terms of the grammar. This means compressing the sample and making sure that the grammar which is derived from the sample is counted as part of the `cost' of compression.

In the case of computer design, we are aiming to compress the system including both the `core' of the computer and all its software. Very roughly, we may equate C in Figure \ref{CONVENTIONAL-AND-SP-COMPUTERS} with the grammar in grammar induction and we may equate S in computers with the data in grammar induction. In the same way that an increase in the size of a grammar can mean an overall decrease in the size of grammar and data taken together, an increase in the size of C can mean an overall reduction in (C + S), as suggested in Figure \ref{CONVENTIONAL-AND-SP-COMPUTERS}.

\subsection{Extending established ideas}

The principles illustrated in Figure \ref{CONVENTIONAL-AND-SP-COMPUTERS} are not new. The same kinds of principle provide the motivation for the creation of database management systems or expert system shells. In this type of application, and others, a system is created which can be used as the foundation for a range of applications of one kind, thus saving the need to reprogram the basic mechanisms in each application. In accordance with remarks above about cost and quality, the use of database management systems and expert system shells can mean dramatic savings in costs and gains in quality compared with programming each application individually.

What is different about the present proposals is that their anticipated scope is much wider than the scope of a database management system or of an expert system shell. Evidence of the possible scope of the system is provided in the sources cited in Section \ref{SCOPE}.

\subsection{Design of an SP computer}

Given current technology, the most straightforward way to create an SP computer is to program it as a software virtual machine running on a conventional computer. Although the SP61 model has been created in this way as a test-bed and demonstrator for theoretical concepts, it is also, in effect, a realisation of an SP computer.

On longer time-scales, creating an SP computer in this way is almost certainly not the best way to do it. Reasons why this should be so are considered in the next two subsections.

\subsubsection{New hardware needed}

In the `core' of a conventional computer (as in the basic mechanisms of the UTM and PCS), mechanisms for searching and matching are not programmed. In the case of the UTM and the PCS, the mechanisms are part of the definition. In digital computers, notwithstanding the existence of `microcode', core mechanisms are, in general, provided by dedicated hardware like the logic circuits mentioned earlier for accessing an item in memory by means of its address. 

In a similar way, it is not necessary for the search mechanisms in an SP computer to be programmed. Creating these mechanisms by programming a conventional computer is a bit like cleaning the bathroom with a toothbrush (as Goldie Hawn was required to do in ``Private Benjamin''). It can be done but the basic tools are clearly not designed for the job. In the long run, it is likely that hardware techniques may be found which meet the requirements of the system more directly.

New hardware does not necessarily mean new {\it electronic} hardware. There is potential for flexible pattern-matching using light with high levels of parallelism (see next), especially since light rays can cross each other without interfering with each other. As noted previously (Section \ref{DNA-COMPUTING}), there is also potential for searching huge pattern-matching search spaces in parallel using chemicals in solution (or other form), as, for example, in DNA computing.

\subsubsection{Parallel processing}

Although the estimated time complexity and space complexity of the current SP model are good, the ICMAUS mechanisms are intrinsically `hungry' for computing power unless they are constrained to work like a conventional computer.
One possible solution to this problem is the creation of an SP system as a software virtual machine running on a high-parallel computer comprising many (cheap) conventional processors. This approach is perhaps more promising than a software virtual machine on a single processor. Returning to the toothbrush analogy, it would be like a cleaning device comprising thousands of toothbrushes - conceivably quite effective.

Alternatives, as before, include new devices based on electronics, light or chemistry, or some combination of those mechanisms.

\section{Conclusion}\label{CONCLUSION}

This article has presented the `SP' concepts in outline and has tried to show how established models of computing may be interpreted in terms of information compression by multiple alignment, unification and search. It has also been argued that established models may be augmented to exploit more fully the potential functionality of the ICMAUS mechanisms.

Although this augmentation would increase the size and complexity of the core computational process, research to date suggests that there could be a reduction in the size and complexity of software - and this could mean an overall saving in the size and complexity of the combination of core system and software. In terms of practicalities, less software should mean lower costs and fewer errors in working systems. There should also be better integration across diverse applications.

In accordance with `Occam's razor' (and, indeed, with the MLE principle incorporated in the SP theory), a good theory should combine `Simplicity' with explanatory or descriptive `Power'. Established models of computing are very simple but they leave many problems unsolved. The conceptual framework described in this article seems to provide an interpretation for established models. At the same time, it suggests how these models may, with advantage, be enlarged. The resulting reduction in Simplicity may be more than offset by an increase in descriptive and explanatory Power.

\section*{Acknowledgements}

I am grateful to one of the referees of this article for alerting me to the relevance of the rule-based programming paradigm, probabilistic parallel programming based on multiset transformation, classifier systems and the TREAT algorithm, and for helpful comments on the article. Thanks also to the second referee for suggesting DNA computing for inclusion in the article. I am grateful also to Professor Tim Porter of the School of Mathematics, University of Wales, Bangor, for very helpful comments on earlier drafts of this article and discussion of related issues. Many thanks also to Dr Chris Wensley in the same School for useful discussions from time to time of related issues.

\appendix
\section{Definitions of concepts}\label{APPENDIX}

\subsection{Basic concepts}

\subsubsection*{Definition 1}

A {\bf symbol} is some kind of mark which can be compared with any other symbol. In the context of pattern matching, a symbol is the smallest unit which can participate in matching: a symbol can be compared (matched) only with another single symbol and the result of matching is either that the two symbols are the same or that they are different. No other result is permitted.

An important feature of the concept of a {\it symbol}, as it is used in this research, is that, with respect to the way it behaves within the system, it has {\it \bf no hidden meaning}. In this research, a symbol is {\it nothing but} a primitive mark which can be discriminated in a yes/no manner from other symbols. There are no symbols like the symbols in an arithmetic function (e.g., `6', `22', `+', `-', `$\times$', `/', `(`, `)' etc), each of which has a meaning for the system which is not directly visible.

Any symbol may have a meaning but that meaning must take the form of one or more other symbols which are associated with the given symbol and are explicit and visible within the structure of symbols and patterns.

Although symbols in the proposed framework have no hidden meaning having an effect on how the symbol behaves in the system, they can and often do have a meaning for the human reader.

If two symbols match, we say that they belong to the same {\bf symbol type}. In any system which contains symbols, we normally recognise an {\bf alphabet} of symbol types such that every symbol in the system belongs in one and only one of the symbol types in the alphabet, and every symbol type is represented at least once in the system.

A positive match between two symbols is termed a {\bf hit}. In an alignment of two more {\it patterns}, one or more unmatched symbols within one pattern between two hits or between one hit and the end of the pattern is a {\bf gap}.

\subsubsection*{Definition 2}

A {\bf pattern} is an array of symbols in one, two or more dimensions. In this article, one dimensional patterns ({\bf sequences} or {\bf strings} of symbols) are the main focus of attention.

The meaning of the term {\it pattern} includes the meanings of the terms {\it substring} and {\it subsequence}, defined next.

\subsubsection*{Definition 3}

A {\bf substring} is a sequence of symbols of length $n$ within a sequence of length $m$, where $n \leqslant m$ and where the constituent symbols in the substring are contiguous within the sequence which contains the substring.

\subsubsection*{Definition 4}

A {\bf subsequence} is a sequence of symbols of length $n$ within a sequence of length $m$, where $n \leqslant m$ and where the constituent symbols in the subsequence may not be contiguous within the sequence which contains the subsequence. The set of all subsequences of a given sequence includes all the substrings of that sequence.

\subsection{The Post Canonical System in normal form}

\subsubsection*{Definition 5}

A {\bf Post Canonical System} in normal form is a quadruple PCS = (A, I, P, S), where A is a finite set (alphabet) of one or more symbols (which Post calls `letters'), I is an `axiom' or `input' comprising a finite string of one or more symbols (drawn with replacement from A), P is a finite set of one or more `productions' and S is a method of searching for a match between a string of symbols and one or more of the productions.\footnote{The definition of a PCS given here is drawn from the descriptions in \cite{r11} and \cite{r10} except that S has been identified explicitly as an element of the system. In both Post's and Minsky's accounts, the process of searching for matches between strings and substrings is not formally identified as an element of the system and the descriptions which they give of S are informal. The reason for identifying S explicitly in this article is to emphasise what appears to be the key difference between a PCS (or UTM) and generalised forms of the SP system, as discussed in Section \ref{AUGMENTATION}.}

Each production has the form:

\begin{center}
$g\ \$\ \rightarrow\ \$\ h$,
\end{center} 

\noindent where $g$ and $h$ each represent a string of zero or more symbols (drawn with replacement from S), and both instances of `\$' represent a single `variable' which may have a value comprising a string of zero or more symbols.

A PCS (in normal form) operates by searching (using the method S) for one or more exact matches between the leading substrings in I and the string $g$ in one or more of the productions (P). Wherever an exact match is found, the trailing substring (if any) is assigned to the variable, \$. The string which comprises the value of `\$' followed by the string $h$ (which may be termed `output') is then treated as new input, and matches are sought for this input as before. The process terminates if and when no more matches can be found.

A key characteristic of S is that, for any `input' string (i.e., the original input string or one of the strings formed by the system which are processed in the same way), it should be able to find every production where the string $g$ in the production is an exact match for a leading substring in the input string.

\subsubsection*{Comments}

Notice that any search for a match between a symbol string and the leading symbols of productions can yield more than one positive result - and a given computation may develop two or more `strands' with further branching at any stage (see \cite[Chapter 13]{r10}. Thus PCSs can accommodate both `deterministic' and `non-deterministic' styles of computing.

\subsection{The SP System}

\subsubsection*{Definition 6}

An SP System is a quadruple SPS = (A$^\prime$, N, O, S$^\prime$), where A$^\prime$ is a finite set (alphabet) of one or more primitive symbols, N is a pattern of symbols (`New'), O is a finite set of one or more patterns of symbols (`Old'), and S$^\prime$ is a search method, different from and, in some sense, more sophisticated than S in a PCS (see below).

Using S$^\prime$, the system operates by searching for one or more {\it alignments} (defined next), where each alignment contains N and one or more patterns from O and where each alignment promotes the compression of N by unification of matching symbols in the alignment.

\subsubsection*{Definition 7}

In the case of one-dimensional patterns,\footnote{As previously noted, the concept of an alignment can be generalised in a straightforward manner to patterns of two or more dimensions. But no attempt is made here to provide a formal definition for alignments of patterns of two dimensions or higher.} an {\bf alignment} is a two-dimensional array of one or more patterns, each one in a separate {\bf row} in the array. The alignment shows sets of two or more matching symbols by arranging the symbols in each set in a {\bf column} of the array.\footnote{The fact that, in displaying alignments, it can sometimes be convenient to put non-matching symbols in the same column with lines to mark the symbols that do match (as in Figure \ref{DNA}) is not relevant to the abstract definition of an alignment presented here.} In an alignment, as defined in this research:

\begin{itemize}
\item Symbols which are contiguous in a pattern which appears in an alignment, need not occupy contiguous cells in the array.

\item Any one pattern may appear one or more times in an alignment.

\item Where a pattern appears two or more times in an alignment, no symbol in one appearance of the pattern should ever be shown as matching the same symbol in another appearance of the pattern.

\item Any symbol in one pattern may be placed in the same column as any other symbol from the same pattern or another pattern, providing {\it order constraints} are not violated.
\end{itemize}

For any alignment, {\bf order constraints} are preserved if the following statement is always true:

\begin{quote}
For any two rows in the alignment, A and B, and any four symbols, A$_1$ and A$_2$ in A, and B$_1$ and B$_2$ in B, if A$_1$ is in the same column as B$_1$, and if A$_2$ is in the same column as B$_2$, and if A$_2$ in A follows A$_1$ in A, then B$_2$ in B must follow B$_1$ in B. 
\end{quote}

\noindent This condition holds when the two rows contain two different patterns and also when the two rows contain two appearances of one pattern.

\subsubsection*{Definition 8}

A {\bf mismatch} in an alignment occurs when, between to columns in the alignment containing hits, or between one column containing hits and the end of the alignment, there are no other columns containing hits and there are two more columns containing single symbols from two or more different patterns in Old.

\subsubsection*{Definition 9}

The {\bf search space} of an SPS is the set of possible alignments of N and the patterns in O.

\subsubsection*{Definition 10}

An alignment is {\bf equivalent} to one step in the operation of a PCS if:

\begin{itemize}

\item The alignment contains no mismatches.

\item Every symbol in I of the PCS can be mapped to an identical symbol in the pattern N in the alignment.

\item The order of the symbols in I is the same as the order of the matching symbols in N.

\item For the production in the PCS which enters into the given operation, every symbol in that production can be mapped to an identical symbol in a pattern from O which appears in the alignment, and the order of the symbols in the production are the same as the order of the matching symbols in the corresponding pattern.

\item For every positive match between two symbols in the operation of the PCS there is a column in the alignment containing a matching pair of symbols. The order of these columns in the alignment must be the same as the order of the corresponding pairs of symbols in the operation of the PCS.
\end{itemize}


\begin{thebibliography}{xxx}

\bibitem[Adleman 98]{r0} Adleman, L. M.: ``Computing with DNA''; {\it Scientific American}, 279, 2 (1998) 54-61.

\bibitem[Adleman 94]{r0a} Adleman, L. M.: ``Molecular computation of solutions to combinatorial problems''; {\it Science}, 266 (1994) 1021-1024.

\bibitem[Allison and Wallace 94]{r1} Allison, L and Wallace, C. S.:
``The posterior probability distribution of alignments and its
application to parameter estimation of evolutionary trees and to
optimization of multiple alignments''; {\it Journal of Molecular
Evolution}, 39 (1994) 418-430.

\bibitem[Allison {\it et al.} 92]{r1a} Allison, L, Wallace, C. S. and
Yee, C. N.: ``Finite-state models in the alignment of macromolecules'';
{\it Journal of Molecular Evolution}, 35 (1992) 77-89.

\bibitem[Allison and Yee 90]{r2} Allison, L. and Yee, C. N.: ``Minimum
message length encoding and the comparison of macromolecules''; {\it
Bulletin of Mathematical Biology (UK)}, 5, 3 (1990) 431-453.

\bibitem[Baum and Boney 96]{r2a} Baum, E. B. and Boneh, D.: ``Running dynamic programming algorithms on a DNA computer''; {\it Proceedings of the Second Annual Meeting on DNA Based Computers}, held at Princeton University, June 10-12, 1996. DIMACS: {\it Series in Discrete Mathematics and Theoretical Computer Science}, (1996) ISSN 1052-1798,

\bibitem[Chan {\it et al.} 92]{r3} Chan, S. C., Wong, A. K. C. and
Chiu, D. K. Y.: ``A survey of multiple sequence comparison methods'';
{\it Bulletin of Mathematical Biology (UK)}, 54, 4 (1992) 563-598.

\bibitem[Clocksin and Mellish 94]{r4} Clocksin, W. F. and Mellish, C.
S.: {\it Programming in Prolog}, Springer-Verlag, Berlin (1994).

\bibitem[Cover and Thomas 91]{r5} Cover, T. M. and Thomas, J. A.: {\it
Elements of Information Theory}, New York: John Wiley (1991).

\bibitem[Felsenstein 81]{r6} Felsenstein, J.: ``Evolutionary trees from
DNA sequences: a maximum likelihood approach''; {\it Journal of
Molecular Evolution}, 17 (1981) 368-376.

\bibitem[Forrest 91]{r6a} Forrest, S.: {\it Parallelism and Programming
in Classifier Systems}, San Mateo, Ca.: Morgan Kaufman (1991).

\bibitem[Kleene 36]{r7} Kleene, S. C.: ``l-definability and
recursiveness''; {\it Duke Mathematical Journal}, 2 (1936) 340-353.

\bibitem[Krishnamurthy and Krishnamurthy 99]{r7a} Krishnamurthy, V. and
Krishnamurthy, E. V.: ``Rule-based programming paradigm: a formal
basis for biological, chemical and physical computation''; {\it Biosystems},
49 (1999) 205-228.

\bibitem[Li and Vit\'{a}nyi 97]{r8} Li, M. and Vit\'{a}nyi, P.: {\it An
Introduction to Kolmogorov Complexity and Its Applications}, Second
Edition, New York: Springer-Verlag (1997).

\bibitem[Markov and Nagorny 88]{r9} Markov, A. A. and Nagorny, N. M.:
{\it The Theory of Algorithms}, Kluwer, Dordrecht (1988).

\bibitem[Murthy and Krishnamurthy 95]{r9a} Murth, V. K. and Krishnamurthy,
E. V.: ``Probabilistic parallel programming based on multiset transformation'';
{\it Future Generation Computer Systems}, 11 (1995) 283-293.

\bibitem[Minsky 67]{r10} Minsky, M. L.: {\it Computation, Finite and
Infinite Machines}, Prentice Hall, Englewood Cliffs, New Jersey
(1967).

\bibitem[Miranker 90]{r10a} Miranker, D. P.: {\it TREAT: A New and Efficient
Match Algorithm for AI Production Systems}, San Mateo, Ca.: Morgan Kaufman (1990). 

\bibitem[Post 43]{r11} Post, E. L.: ``Formal reductions of the general
combinatorial decision problem''; {\it American Journal of
Mathematics}, 65 (1943) 197-268.

\bibitem[Reichert{\it et al.} 73]{r12} Reichert, T. A., Cohen, D. N.
and Wong, A. K. C.: ``An application of information theory to genetic
mutations and the matching of polypeptide sequences''; {\it Journal of
Theoretical Biology}, 42 (1973) 245-261.

\bibitem[Rissanen 78]{r13} Rissanen, J.: ``Modelling by the
shortest data description''; {\it Automatica}, 14 (1978) 465-471.

\bibitem[Rosser 84]{r14} Rosser, J. B.: ``Highlights of the history of
the lamda-calculus''; {\it Annals of the History of Computing (USA)},
6, 4 (1984) 337-349.

\bibitem[Roytberg 92]{r15} Roytberg, M. A.: ``A search for common
patterns in many sequences''; {\it Cabios}, 8, 1 (1992) 57-64.

\bibitem[Solomonoff 64]{r16} Solomonoff, R. A.: ``Formal theory of
inductive inference, Parts I and II''; {\it Journal of Information and
Control}, 7 (1964) 1-22 and 224-254.

\bibitem[Solomonoff 97]{r17} Solomonoff, R. A.: ``The discovery of
Algorithmic Probability''; {\it Journal of Computer and System
Science}, 55, 1 (1997) 73-88.

\bibitem[Storer 88]{r18} Storer, J. A.: {\it Data Compression, Methods
and Theory}, Computer Science Press, Rockville, Maryland (1988).

\bibitem[Turing 36-7]{r19} Turing, A. M.: ``On computable numbers with
an application to the entscheidungsproblem''; {\it Proceedings of the
London Mathematical Society}, series 2, 42 (1936-7) 230-265 and
544-546.

\bibitem[Turing 50]{r20} Turing, A. M.: ``Computing machinery and
intelligence''; {\it Mind}, 59 (1950) 433-460.

\bibitem[Wallace and Boulton 68]{r21} Wallace, C. S. and Boulton, D.
M.: ``An information measure of classification''; {\it Computer
Journal}, 11, 2 (1968) 185-194.

\bibitem[Wolff 99]{r21a} Wolff, J. G.: ``Probabilistic reasoning as
information compression by multiple alignment, unification and search:
an introduction and overview''; {\it Journal of Universal Computer
Science}, 5, 7 (1999) 417-472. A copy may be obtained from:
http://www.iicm.edu/jucs\_5\_7.

\bibitem[Wolff 98a]{r22} Wolff, J. G.: ``Probabilistic reasoning as
information compression by multiple alignment, unification and search
(I): introduction''; SEECS Report, 1998. A copy may be obtained from:
http://www.iicm.edu/wolff/1998a or from:
http://www.sees.bangor.ac.uk/$\sim$gerry/sp\_summary.html\#PrbRsI.

\bibitem[Wolff 98b]{r23} Wolff, J. G.: ``Probabilistic reasoning as ICMAUS (II): calculation of probabilities, best-match pattern recognition and information retrieval, and reasoning with networks, trees and rules''; SEECS Report 1998. A copy may be obtained from: http://www.iicm.edu/wolff/1998b or from: http://www.sees.bangor.ac.uk/$\sim$gerry/sp\_summary.html\#PrbRsII.

\bibitem[Wolff 98c]{r24} Wolff, J. G.: ``Probabilistic reasoning as ICMAUS (III): hypothetical reasoning, geometric analogies, default values, nonmonotonic reasoning, and modelling `explaining away'''; SEECS Report 1998. A copy may be obtained from: http://www.iicm.edu/wolff/1998c \linebreak or from: http://www.sees.bangor.ac.uk/$\sim$gerry/sp\_summary.html\#PrbRsIII.

\bibitem[Wolff 98d1]{r25} Wolff, J. G.: ``Parsing as information
compression by multiple alignment, unification and search''; SEECS
Report, 1998. A copy may be obtained from: http://www.iicm.edu/wolff/1998d1. This article is based on \cite{r25a} and \cite{r25b}.

\bibitem[Wolff 98d2]{r25a} \begin{flushleft} Wolff, J. G.: ``Parsing as
information compression by multiple alignment, unification and search:
SP52''; SEECS Report, 1998. A copy may be obtained from: http://www.iicm.edu/wolff/1998d2 or from: http://www.sees.bangor.ac.uk/$\sim$gerry/sp\_summary.html\#parsing. \end{flushleft}

\bibitem[Wolff 98d3]{r25b} \begin{flushleft} Wolff, J. G.: ``Parsing as
information compression by multiple alignment, unification and search:
examples''; SEECS Report, 1998.  A copy may be obtained from:
http://www.iicm.edu/wolff/1998d3 or from:
http://www.sees.bangor.ac.uk/$\sim$gerry/sp\_summary.html\#parsing.
\end{flushleft}

\bibitem[Wolff 97]{r26} Wolff, J. G.: ``Causality, statistical learning
and multiple alignment''; Paper presented at the UNICOM Seminar and
Tutorial on Causal Models and Statistical Learning, London, March
1997.

\bibitem[Wolff 96]{r27} Wolff, J. G.: ``Learning and reasoning as
information compression by multiple alignment, unification and
search''; In: A. Gammerman (ed.), {\it Computational Learning and
Probabilistic Reasoning}, Wiley, Chichester (1996) 67-83. An earlier
version was presented at Applied Decision Technologies '95, Brunel
University, April 1995 (Proceedings of Stream 1, {\it Computational
Learning and Probabilistic Reasoning} 223-236).

\bibitem[Wolff 95a]{r28} Wolff, J. G.: ``Computing as compression: an
overview of the SP theory and system''; {\it New Generation Computing},
13 (1995) 187-214.

\bibitem[Wolff 95b]{r29} Wolff, J. G.: ``Computing as compression:
SP20''; {\it New Generation Computing} 13 (1995) 215-241.

\bibitem[Wolff 94a]{r30} Wolff, J. G.: ``A scaleable technique for
best-match retrieval of sequential information using metrics-guided
search''; {\it Journal of Information Science}, 20, 1 (1994a) 16-28.

\bibitem[Wolff 94b]{r31} Wolff, J. G.: ``Towards a new concept of
software''; {\it Software Engineering Journal}, 9, 1 (1994b) 27-38.

\bibitem[Wolff 94c]{r32} Wolff, J. G.: ``Computing and information
compression: a reply''; {\it AI Communications}, 7, 3/4 (1994c)
203-219.

\bibitem[Wolff 93]{r33} Wolff, J. G.: ``Computing, cognition and
information compression''; {\it AI Communications}, 6, 2 (1993)
107-127.

\bibitem[Wolff 91]{r34} Wolff, J. G.: {\it Towards a Theory of
Cognition and Computing}, Ellis Horwood, Chichester (1991).

\bibitem[Wolff 90]{r35} Wolff, J. G.: ``Simplicity and power: some
unifying ideas in computing''; {\it Computer Journal}, 33, 6 (1990)
518-534.

\bibitem[Wolff 88]{r36} Wolff, J. G.: ``Learning syntax and meanings
through optimization and distributional analysis''; In Y. Levy, I. M.
Schlesinger and M. D. S. Braine (Eds.), {\it Categories and Processes
in Language Acquisition}, Lawrence Erlbaum, Hillsdale, NJ (1988).
Reprinted in Chapter 2 of \cite{r34}.

\bibitem[Wolff 82]{r37} Wolff, J. G.: ``Language acquisition, data
compression and generalization''; {\it Language and Communication}, 2
(1982) 57-89. Reprinted in Chapter 3 of \cite{r34}.
\end{thebibliography}
\end{document}